\documentclass[final,3p,times]{elsarticle}

\usepackage{amsmath,amsfonts,bm}

\def\eqref#1{equation~\ref{#1}}

\def\1{\bm{1}}

\DeclareMathAlphabet{\mathsfit}{\encodingdefault}{\sfdefault}{m}{sl}
\SetMathAlphabet{\mathsfit}{bold}{\encodingdefault}{\sfdefault}{bx}{n}

\DeclareMathOperator*{\argmin}{arg\,min}

\usepackage{amsmath}
\usepackage{graphicx}
\usepackage{amsfonts}      
\usepackage{nicefrac}       
\usepackage{microtype}    
\usepackage{xcolor}        
\usepackage{booktabs,adjustbox}   
\usepackage{amssymb}
\usepackage{tabularx}
\usepackage{subcaption}
\usepackage[table]{xcolor}

\usepackage{hyperref}
\usepackage{url}

\renewcommand{\v}[1]{{\boldsymbol{\mathbf{#1}}}}
\newcommand{\state}{\v{s}}
\newcommand{\statederiv}{\v{s}^\prime}
\newcommand{\genstate}{{\state^\dagger}}
\newcommand\mask{m}
\renewcommand{\argmin}{\operatorname*{argmin}}

\graphicspath{{img/},{img2/}}

\journal{under review}

\begin{document}

\begin{frontmatter}

\title{Object-centric proto-symbolic behavioural reasoning from pixels}

\author[inst1]{Ruben van Bergen}

\affiliation[inst1]{organization={Donders Institute, Radboud University},
            city={Nijmegen},
            country={The Netherlands}}

\author[inst1]{Justus Hübotter}

\author[inst2]{Alma Lago}

\author[inst1,inst2]{Pablo Lanillos*}

\affiliation[inst2]{organization={Cajal Neuroscience Center, Spanish National Research Council},
            city={Madrid},
            country={Spain}}

\begin{abstract}

Autonomous intelligent agents must bridge computational challenges at disparate levels of abstraction, from the low-level spaces of sensory input and motor commands to the high-level domain of abstract reasoning and planning. A key question in designing such agents is how best to instantiate the representational space that will interface between these two levels---ideally without requiring supervision in the form of expensive data annotations. These objectives can be efficiently achieved by representing the world in terms of objects (grounded in perception and action). In this work, we present a novel, brain-inspired, deep-learning architecture that learns from pixels to interpret, control, and reason about its environment, using object-centric representations. We show the utility of our approach through tasks in synthetic environments that require a combination of (high-level) logical reasoning and (low-level) continuous control. Results show that the agent can learn emergent conditional behavioural reasoning, such as $(A \to B) \land (\neg A \to C)$, as well as logical composition $(A \to B) \land (A \to C) \vdash A \to (B \land C)$ and XOR operations, and successfully controls its environment to satisfy objectives deduced from these logical rules. The agent can adapt online to unexpected changes in its environment and is robust to mild violations of its world model, thanks to dynamic internal desired goal generation. While the present results are limited to synthetic settings (2D and 3D activated versions of dSprites), which fall short of real-world levels of complexity, the proposed architecture shows how to manipulate grounded object representations, as a key inductive bias for unsupervised learning, to enable behavioral reasoning.
\end{abstract}


\end{frontmatter}

\section{Introduction}
\label{sec:intro}

A one-year old infant, before language is expressed~\cite{cowley2007human}, learns the efferent-afferent patterns of sensory stimulation and motor commands to a neural representation of the environment.
The pathway from the sensorium, to abstract thought, and back to the minutiae of the sensorimotor domain defines the feats of higher-order cognition and structures the different levels of abstraction at which intelligent agents must operate, when they interact with the environment~\cite{gibney2024ai}. How sub-symbolic computations transform into higher-level cognitive representations of structures is far from being understood~\citep{piantadosi2021computational}. However, there is common accepted idea that suggest that there are intermediate representations that bridge cognitive reasoning and behaviour~\citep{battaglia2013simulation}. Hence, in the design of artificial agents, a key challenge is first to find a representational space that provides an effective interface between these disparate demands, as well as a mechanism that makes proper use of this representation to interact with the environment and produce behaviour. In this sense, we approach reasoning from a behavioral perspective where the output of a reasoning process is always an action that physically interacts with the environment.

To address these challenges in a single, comprehensive system, here, we introduce a novel, brain-inspired neural network architecture that spans the domains of cognitive reasoning, perceptual inference, planning and continuous control. We follow both the emergentist and probabilistic approach to cognition, where the representational format, and the machinery to transform sensory observations into this space, is learned unsupervised but allows reasoning as an inference process. This also avoids the dependency on ground-truth labels furnished by human annotators, as these are highly labor-intensive to produce, especially when labels must cover all the relevant variables in a scene. The proposed architecture leverages the core inductive bias that the environment can be partitioned into discrete entities, or \textit{objects}, that obey many useful symmetries and invariances~\cite{greff2020binding}. Object-based representations and reasoning are a key tenet of perception and cognition in humans~\citep{Peters2021}. Objects constitute the environment's separable, movable parts, as well as the logical units of reasoning and planning. They naturally take on properties of symbols, as different objects obey the same laws of physics and possess similar or analogous properties~\cite{griffiths2010probabilistic}. At the same time, each object representation is tethered to the sensorimotor domain via explicit attention maps. Objects, thus, are an ideal level of abstraction to interface between the disparate levels of computation that we require. We show the promise of our approach by evaluating it in tasks that require both high-level reasoning and continuous control in synthetically generated environments.

\section{Related work, challenges and contribution}
\label{sec:intro:contrib}

Structured representation learning appeared as a powerful way to introduce inductive biases and scale artificial intelligence to high-level cognition~\citep{goyal2022inductive} and exploit symmetries and invariances that can be leveraged by decomposing scenes into their natural constituents~\citep{eslami2016attend}. Particularly, object-centric representations provide a natural interface between bottom-up (e.g., emergentist, connectionist) and top-down (e.g., graph based, probabilistic) approaches~\citep{griffiths2010probabilistic}. We summarize the relevant related works that influenced the proposed architecture classified in four topics\footnote{Some of the works may have overlapping features in other topics.}: scene understanding (e.g., visual segmentation and prediction), physics-informed and simulation-based methods, robotics and reasoning (e.g., object relations). We analyse these works from the prism of behavioural reasoning. Furthermore, Table~\ref{tab:SOTA} details the object-centric relevant work depending on the input and output: (A) Object control with full access to the object positions. (B) Object-centric image and video prediction and (C) Image-based object-centric control and planning.

\paragraph{Visual segmentation and prediction}
Visual segmentation methods have reached maturity with high accuracy on image segmentation and prediction. {While non-object-centric powerful video prediction methods, such as the PredRNN~\cite{wang2022predrnn}, can extract object shapes and are usually used in object-based benchmarking. Here, we narrow down to object-centric solutions in both static images~\citep{Greff2019a_Iodine} and videos~\citep{elsayed2022savi++}. From IODINE~\citep{Greff2019a_Iodine} that used iterative amortized inference, recent methods evolved to slot-attention architectures~\citep{locatello2020SlotAttention} and diffusion approaches~\citep{jiang2023object}. These methods, which can track 2D and 3D objects in cluttered scenes~\cite{traub2022learning}, capture object collisions and discover new objects~\cite{yu2022unsupervised}, are designed for perception and not for reasoning about, or controlling the segmented objects. 
Connecting visual prediction algorithms to a controller usually not provide the best performance, as the features and the transition or dynamics learnt are not coupled with the action information. For instance, in world models prediction like both activated and passive pretrained options are provided~\cite{assran2025v}. Our proposed model learns a transition model in the latent space through acting in the environment, tightening the object-centric representation to the planned behaviour.

\paragraph{Physics-informed and simulation-based methods}
These approaches have also shown a great potential in handling complex non-linear interactions. These are strongly influenced by the simulation approach to cognition~\cite{battaglia2013simulation}. For instance, interaction~\cite{battaglia2016interaction} and propagation~\cite{li2019propagation} networks are able to understand complex scene dynamics with multiple objects and even control them to reach a desired known state. Unfortunately, these methods require full observation of the objects states. Approaches that can work directly on pixels make use of ground-truth segmentation masks~\cite{piloto2022intuitive} or Visual interaction networks~\cite{watters2017visual}. While these approaches have some kind of physical ``reasoning" a full-fledged architecture that performs 1st order logic reasoning from pixels and transforms to behaviour (control) is still not fully accomplished.

\paragraph{Robotics}
These works focus on moving the objects meaningfully and usually have full access to the objects states (e.g. \citep{sancaktar2022curious}). OP3~\citep{veerapaneni2020entity} is an outstanding exception. However, it has two key limitations. First, it is restricted to a discrete action space of picking up and placing objects, with no continuous control. Second, it has no ability to learn tasks---it can only plan actions towards objectives specified \textit{ad hoc} by means of a goal image that shows exactly what the scene in question ought to look like. Novel approaches use ReFs to describe the 3D representation of the object and control is solved by RRTs in the latent prediction dynamics and Model Predictive Control (MPC) to execute the actions~\citep{driess2023learning}. Still, ground truth objects mask are used for training and inference. Some methods have overcome this constraint through entity-based segmentation---as a simplification of object-centric representation that uses the center of the object as the location of the entity---and RL to overcome with the generation of meaningful behaviours~\cite{haramati2024entity}. Unfortunately, this restricts the shape of objects to ``point mass" like. Besides, there are object-centric approaches that, instead of focusing on the interactive behaviour between the agent and the objects, focus on view-point matching~\cite{van2024object} and agent navigation~\cite{huo2023geovln}.  In terms of architecture, according to the classification provided in \cite{garrido2024learning}, our approach would lie into the \textit{JEPA} models, where prediction happens in the latent space and includes explicit control. While V-JEPA 2-AC~\cite{assran2025v} uses a more powerful embedding encoding than our dynamic iterative variational inference perceptual backbone, they still suffer the same limitation as in other goal-image-conditioned manipulation: ``the optimization target assumes that we have access to visual goals”. We address this limitation in OBR by including a preference network, allowing behavioural 1st order logic. Another important factor is the inference time. \cite{assran2025v} reports an inference time of 16 seconds per action plan (using cross-entropy optimization), and this is not still suitable for a real-world setting. While our current architecture does not generalize to realistic images, action computation is computed in closed form, achieving fast inference times (less than one second). Recent architectures combining slot-attention and model-based RL~\cite{sold2025mosbach} are also providing suitable inference times but their training time is comparably much higher.

\paragraph{Reasoning}
Leaving out the literature on visual scene understanding with language, reasoning with object-centric representations as neurosymbols is, while the most promising, the least investigated area~\citep{assouel2022object}. There are just a few object-centric studies on visual reasoning in static images~\cite{webb2024systematic}. Furthermore, there are relevant works from Large Language Models research (e.g., \cite{driess2023palm}), but they do not adhere to the grounding paradigm proposed, where reasoning should appear as an emergent property of learning a world physical model~\cite{taniguchi2023world}. For instance in~\cite{ahn2022can} the reasoning capabilities are much more expressive than our proposed approach but they use pre-trained behaviours, which are conditioned on the language generated by the LLM. Conversely, our architecture learns the world dynamics and interaction through unsupervised learning, harnessing the construction of the proto-symbols while interacting. This does not prevent the possibility of connecting proposed architecture to an LLM similarly to~\cite{xu2023llms}, but through the preference network, which is already grounded.

\begin{table}
\caption{Relevant approaches related to object centric behavior generation. We categorized the methods into three blocks: (A) Object control with full access to the object positions, (B) Object-based image and video prediction---while they do not encode actions, they provide the architectures for high-dimensional object-centric control solutions---and (C) Image-based object-centric control and planning}
\label{tab:SOTA}
\renewcommand{\arraystretch}{1.2}
\resizebox{\textwidth}{!}{%
\begin{tabular}{@{}l p{3.5cm} l p{2.5cm} p{2.5cm} p{2.5cm} p{3.5cm} p{6.5cm}@{}}
\toprule
\textbf{Methods} & \textbf{Input/Action} & \textbf{Perception} & \textbf{Transition} & \textbf{Control} & \textbf{Goal input} & \textbf{Objective} & \textbf{Strengths / Limitations} \\ 
\midrule
\rowcolor{gray!15} \multicolumn{8}{l}{\textit{A. Object-based prediction and control with full access to positions}} \\ 
Interaction Net~\cite{battaglia2016interaction} & Obj. pos / - & - & graph encoding & No & - & - & Predicts non-linear interactions (visual encoder version exists). \\
 \hline
    Propagation Net\cite{li2019propagation}                                              & Obj. pos / Position or velocity                       & -                         &  graph encoding                      & SGD +MLP          & Desired obj. positions                             & Chanfer distance     & Allows control of complex non-linear interactions. / needs ground truth object positions.                                                 \\ \hline
    CEE-US~\cite{sancaktar2022curious}                                                         & Obj. pos / End-effector pos (2D plane)                &   -                     & Graph Neural Networks                      & RL                & Desired obj. positions                             & Max E[r]             & Incorporates curiosity drivers. / needs ground truth object positions.                                                                    \\ \hline

\\
    \rowcolor{gray!15} \multicolumn{8}{l}{\textit{B. Object-based image and video prediction. The dynamics are inferred, but no agent action is involved.}} \\

    IODINE~\cite{Greff2019a_Iodine}                                                            & Image/-                                               & IAI~\cite{Marino2018}                   & No                     & -                 & -                                                  & -                    & Perception works like a filter with the refinement network. / Only static images.                                                         \\ \hline
    SlotAttention~\cite{locatello2020SlotAttention}                                                   & Image/-                                               & Slot Attention            & GRU                    & -                 & -                                                  & -                    & Reduced inference time for object representation.                                                                                            \\ \hline
    SAVI~\cite{kipfconditional_SAVi,elsayed2022savi++}                                                          & Video/-                                               & Slot Attention            & GRU                    & -                 & -                                                  & -                    & Works for realistic videos / Tends to smooth the shapes into blobs and does not incorporate control.                                                                                   \\ \hline
    LOCI~\cite{traub2023learning}                                                           & Video/-                                               & Autoencoder               & Gatel0rd\cite{gumbsch2021sparsely}            & -                 & -                                                  & -                    & Uses explicit 2D positional encodings as Gaussians.                                                                                                        \\
    \\
    \rowcolor{gray!15} \multicolumn{8}{l}{\textit{C. Object-centric control. The agent can perform actions that change the objects state.}} \\
    OP3~\cite{veerapaneni2020entity}                                                              & Img seq. / Discrete (primitives)                      & IAI\cite{Marino2018}                   & Approximate dynamics network                      & Cross-Entropy Optimization & Encoded goal image                                 & Min latent distance. & Coupled action and perception / only allows discrete  actions                                                                              \\ \hline
    SMORL~\cite{zadaianchuk2020self}                                                            & Img seq. / End-effector pos (2D plane)                & SCALOR\cite{jiang2019scalor}                & Attention head         & SAC+HER           & Encoded goal image                                 &  Distance between object position and goal if they are similar in features                   & Allows RL framework integration / Scales poorly to the  number of objects.                                                                 \\ \hline
    Entity RL\cite{veerapaneni2020entity}                                                        & Img seq./ EE joint position                           & Deep Latent Particles~\cite{daniel2022unsupervised} & Transformer            & TD3+HER           & Reward (distance between state and goal particle). & Max E[r]             & Scales linearly with the number of objects / cannot encode  complex object properties (e.g., shapes).                                      \\ \hline
   SOLD~\cite{sold2025mosbach}                                                        & Image seq./ Continuous positions                           & SAVI~\cite{kipfconditional_SAVi} & Transformer~\cite{villar2023object}            &  Dreamer-like RL  & Task reward & Max E[r] + Entropy regularizer            & Beats monolitic approaches in relational reasoning / Very costly to train and shapes are not properly captured. \\ \hline
    \textbf{OBR} (proposed)                                                               & Img seq. / Continuous (3D acc. vectors)               & IAI~\cite{Marino2018}                   & Latent linear dynamics & Closed-form Expected Free Energy~\cite{millidge2021whence}  & Preference network (set-structured MLP)            & composite ELBO loss & Allows 1st order proto-object behavioural reasoning and fast action computation. / Objects' interactions are not explicitly modelled.                      \\ \hline

    \end{tabular}
    }
\end{table}

\subsection{Current challenges}
There are three major challenges in unsupervised learning with object-centric representations: $i)$ working with complex naturalistic scene images~\cite{jiang2023object}, $ii)$ the generation of meaningful physical actions (i.e., interacting through continuous control) when the input is high-dimensional (i.e., image) and the environment is partially observable and $iii)$ behavioural reasoning, where these actions adhere to a high-order goal, such as logics. This work focuses on the latter. We show that simple conditional reasoning problems, such as, ``if there is a heart move squares to the left, otherwise move squares to the right" ($(A \to B) \land (\neg A \to C)$) is already unsolvable for non-object-centric representations SOTA algorithms---see Sec.~\ref{sec:results}. 
An underlying problem for the challenges is to achieve high performance in world state dynamic estimation (world model learning), but allowing adaptation and generalization. We designed the object-centric backbone architecture to incorporate adaptation, following the free energy principle theory of neural information processing~\cite{friston2010free,lanillos2021active}, where the brain estimates the world by continuously approximating its internal model, learned by experience, to the real world through approximate Bayesian computation. 

Finally, another relevant challenge, which is not addressed in this work, is to deal with complex non-linearities of physical contact, such as collisions. Physics-based simulation-based approaches~\cite{piloto2022intuitive} and Deep Reinforcement Learning (RL)~\cite{haramati2024entity,sold2025mosbach} variants are showing the best results.

\subsection{Contribution}
This work provides a neural network architecture (Fig. \ref{fig:architecture}) that can learn 1st order conditional behavioural reasoning where proto-symbols\footnote{We use the term proto-symbols as in visual segmentation we use for proto-objects~\cite{walther2006modeling}, thus describing candidates for symbols and objects.} take the form of object-centic representations and perception, control and preferences (desired internal state of the agent) are dynamical processes computed through approximate Bayesian inference, in an unsupervised learning scheme from pixels. Our approach draws inspiration from natural intelligence, where the agent's function is to generate behaviours through reasoning conditioned on perceptual cues by optimizing a single quantity: the variational free energy (or evidence lower bound)~\cite{friston2010free}. To this end, we combine bottom-up and top-down object-centric approaches~\citep{rasouli2020attention,van2023object} following the emergentist approach (with unsupervised learning from pixels) but allowing probabilistic inference on the structured representation. This permits the agent to learn: $i)$ what is an object (scene understanding), $ii)$ how to mentally manipulate it through object-centric rules (reasoning) and $iii)$ generate physical behavior (i.e., continuous control). The agent can only perceive through visual input (2D image projection) of a synthetic world composed of 2D or 3D objects (see Fig. \ref{fig:results:intro}a for schematic). It can interact with the environment by applying forces to locations in the image, which corresponds to forces in the 2D/3D world. This architecture allows the agent to learn complex conditional rules, such as $(A \to B \land C) \land (\neg A \to D \land E)$, perform logical composition $(A \to B) \land (A \to C) \vdash A \to (B \land C)$ by learning two different rules separately but then they emerge combined during execution, and XOR operations, such as $(A \lor B) \to C) \land (A \land B \to D)$. Once the agent learns the rules, reasoning transforms into a behavioural response that changes the environment towards its internal preference. Furthermore, the agent, using the proposed architecture, shows online adaptation to unexpected changes in its environment, is robust to mild violations of its world model, and invariant to the number of objects.

\section{Proposed neural network model}
\label{sec:model}
The Object-centric Behavioural Reasoner (OBR) proposed, depicted in Fig. \ref{fig:architecture}, is an object-centric deep learning architecture consisting of two interconnected modules: $i$) perceptual inference, in charge of learning object representations (properties and dynamics) and producing top-down attention and $ii$) action inference, with the reasoning and control, in charge of learning proto-symbolic rules at the level of object representations and transforming them into meaningful behaviors (i.e., control commands). 

\begin{figure}[ht!]
	\centering
        \includegraphics[width=0.9\textwidth]{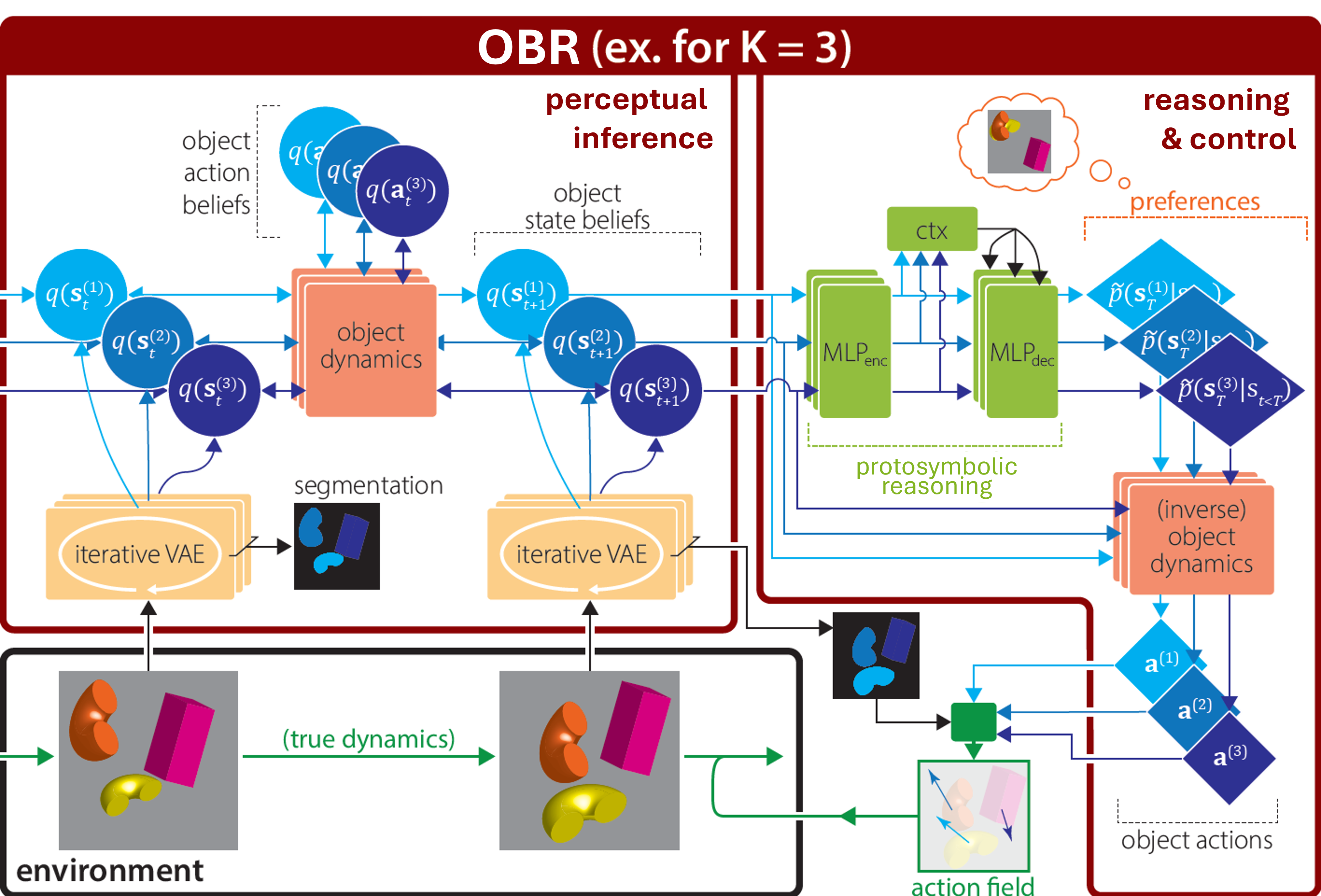}
        \caption{\textbf{OBR model architecture}. Illustration of an example instance of OBR perceptual and reasoning interconnected modules with $K=3$ slots, and an inference window of two time points. The perceptual module exploits iterative amortized inference (through a refinement network). The action module reasons what is the internal state preference (through online goal imagination) and generates the continuous control actions accordingly to obtain the desired state in the real world (through the minimization of the variational free energy). OBR uses object-centric representations, dynamics, reasoning, and control. For clarity, the computation of object action beliefs is not included here (see Appendix \ref{sec:appendix}).} 
	\label{fig:architecture}
\end{figure}

\subsection{Perceptual inference and generative model}

The perception module infers the object state and action beliefs from incoming sensory data, e.g., one RGB image (frame) per timepoint, and the agent's own motor efferents (see Appendix \ref{sec:network-architectures}). 
\paragraph{Object-centric representation} We define the $k$-th object at time $t$ as a state vector $\genstate^{(k)}_t$. This vector is expressed in second-order generalized coordinates - that is, it is a concatenation of the current state and its derivative: ${\genstate}^{(k)}_t = \begin{bmatrix} {\state^{{(k)}}_t}^T, \ {{\statederiv}^{(k)}_t}^T \end{bmatrix}^T$. Objects are influenced by actions from the agent. The action (or action-effect) on object $k$ at time $t$ is denoted by $\v{a}^{(k)}_{t}$. OBR represents its knowledge about object states and actions through variational beliefs $\{q(\genstate^{(k)}_t )\}$ and $\{q(\v{a}^{(k)}_t )\}$, which are mean-field Gaussian distributions parameterized by a mean and variance for each latent dimension.

To learn object-centric representations and perform inference from pixels, we use a set of $K$ weight-sharing iterative variational autoencoders (itVAEs)~\citep{Marino2018, Greff2019a_Iodine}, where $K$ is the number of object representations to be inferred. These itVAEs together invert a generative model in which image pixels are drawn from a Gaussian mixture distribution:
\begin{gather}
    p(o_i|\{ \v{s}^{(k)}\}_{k\in1:K}) = \sum_k \hat{\mask}_{ik}  \mathcal{N}\left(g_{i}(\v{s}^{(k)}), \sigma_o^2 \right) \\
    \hat{\mask}_{ik} = p(\mask_i| \{ \v{s}^{(k)} \}_{k\in1:K}) = \text{Cat}\left(\text{Softmax}\left(\{\pi_i( \v{s}^{(k)} )\}_{k\in1:K}\right)\right)
\end{gather}
where $o_i$ is the value of the $i$-th image pixel, $g_i(\bullet)$ is a decoder function that translates an object state to a predicted mean value at pixel $i$, $\sigma_o^2$ is the variability of pixels around their mean values, and $\pi_i(\bullet)$ maps an object state to a log-probability at pixel $i$, which defines the predicted probability $\hat{\mask}_{ik}$ that the pixel belongs to that object (its segmentation mask). We implement $g_i(\bullet)$ and $\pi_i(\bullet)$ jointly in the decoder of the itVAE, which thus outputs 4 channels per pixel (3 RGB color values + 1 mask logit). Each itVAE thereby predicts its object's subimage and segmentation mask, and a full image prediction can be obtained as a segmentation-weighted superposition of subimages.

\paragraph{Object dynamics} To complete the world model, this per-frame observation model is combined with linearized object-centric dynamics $p(\v{s}^{\dagger(k)}_{t}|\v{s}^{\dagger(k)}_{t-1}, \v{a}^{(k)}_{t-1})$ defined by the following equations:
\begin{align}
    \v{s}_{t}^{\prime(k)} = \v{s}_{t-1}^{\prime(k)} + \v{D}(\v{a}_{t-1}^{(k)}) + \sigma_s\v{\epsilon}_{1_t}, \quad \quad
    \v{s}_{t}^{(k)} = \v{s}_{t-1}^{(k)} + \v{s}_{t}^{\prime(k)} + \sigma_s\v{\epsilon}_{2_t} \label{eq:dynamics2}
\end{align}
where $\v{\epsilon}_{1_t}$ and $\v{\epsilon}_{2_t}$ are noise realizations drawn from a Normal distribution. Note that we are assuming that the latent representation dynamics can be captured with a 2nd-order generalized coordinates with additive noise, but this is not imposed to the environment\footnote{Robotic experiments showed that 2nd order generalized coordinates are enough to track a dynamical system~\citep{bos2022free} depending on the noise source.} The action $\v{a}_t^{(k)}$ on object $k$ at time $t$ is a (2-D or 3-D) vector that specifies the control command (e.g., acceleration) on the object in environment coordinates. Finally, \v{D} is a learned function that transforms the control command to its effect in the model's latent space.

\paragraph{Learning and inference} Inference for each object is performed within a sliding window spanning the present and a number of past time points.
 
Each itVAE infers beliefs for a single object and time point, taking a total of 8 inference iterations to refine these beliefs to their optimal (minimum-ELBO) values. Inference is coupled between itVAEs through information flow between represented objects and between time points in the inference window (see Appendix~\ref{sec:appendix}). OBR optimizes a composite ELBO loss $\mathcal{L}_\text{comp}= \sum_{n=1}^{N_\text{iter}} \frac{n}{N_\text{iter}}\mathcal{L}^{(n)}$ for both learning and inference, where $n$ indexes inference iterations and $\mathcal{L}$ is:
\small
\begin{multline}
    \mathcal{L} = -\sum_{t=0}^{T} \Bigg[ 
    \underbrace{\mathcal{H}\left(q\left(\{\v{s}^{\dagger(k)}_t, \v{a}^{(k)}_t\}\right) \right)}_\text{Complexity}
    + \underbrace{\beta E_{q(\{\v{s}^{(k)}_t\})}[\log p(\v{o}_t|\{ \v{s}^{(k)}_t\})]}_\text{Reconstruction accuracy}  \\
    + \underbrace{\sum_k E_{q(\v{a}^{(k)}_t)}[\log p(\v{a}^{(k)}_t|\v{\Psi}_t)]}_\text{Action inference accuracy}
    + \underbrace{ \sum_k E_{q\left(\v{s}^{\dagger(k)}_{t}, \v{s}^{\dagger(k)}_{t-1}, \v{a}^{(k)}_{t-1}\right)}[\log p(\v{s}^{\dagger(k)}_{t}|\v{s}^{\dagger(k)}_{t-1}, \v{a}^{(k)}_{t-1})]}_\text{Temporal consistency}  \Bigg]
    \label{eq:elbo}
\end{multline}
\normalsize
where $\v{\Psi}_t$ defines the \textit{action field}--- described in the next section--- that maps object-centric actions into to real actions.

\paragraph{Action mapping module}
A key challenge in multi-object environments is that an agent's internal representation of objects has an unknown (and possibly imperfect) correspondence to the objects in the environment. Even if the object features are inferred with perfect accuracy, the order of the objects in the representation is arbitrary. To solve this correspondence problem, we define an action space (e.g., accelerations) that can be placed at pixel locations within the environment's image grid, and objects receive the sum of all accelerations that coincide with their visible pixels. Specifically, we introduce the notion of an \textit{action field}, which transform into real actions if the pixel belongs to a specific object in the environment, $\v{\Psi}=[\v{\psi}_1, ..., \v{\psi}_M]^T$: an $[M \times 2]$ matrix (with $M$ the number of pixels in an image or video frame), such that the $i$-th row in this matrix ($\v{\psi}_i$) specifies the (x,y,z)-acceleration applied at pixel $i$.  The action on the $k$-th object is then given by:
\begin{equation} \label{eq:actionfield}
    \v{a}_t^{(k)} = \sum_i [\mask_i=k]\v{\psi}_i
\end{equation}
where $\mask_i$ is a categorical variable that indicates which object pixel $i$ belongs to
\footnote{Note the use of Iverson-bracket notation; the bracket term is binary and evaluates to 1 iff the expression inside the brackets is true.}. This definition of actions in pixel space provides an unambiguous interface for the agent to interact with its environment.

The action inference module does not incorporate a decoder network, as the quality of the action beliefs is computed by evaluating equation \ref{eq:actionfield} and plugging this into the ELBO loss from equation \ref{eq:elbo}. While this requires some additional sampling operations (see Appendix~\ref{sec:samplingtrick}), no neural network is required for this. This module does include a (shallow) refinement network---detailed in the Appendix~\ref{sec:appendix}. This network takes as input the current variational parameters $\v{\lambda_a}^{(k)}$ (means and variances), their gradients, and the `expected object action', $\sum_i \hat{\mask}_{ik} \v{\psi}_i$. 

\subsection{Reasoning and control}
OBR's generative model allows it to infer the current state of the world, as well as predict future states that would arise as the result of the agent's actions. In a (model-based) RL framework, we could now define rewards and learn a value function on which to base a behavioral policy. Here, we take a slightly different approach, which is seminal in the proposed architecture, based on the influential computational neuroscience framework of Active Inference \citep{parr2022active,lanillos2021active}. It posits that agents, through their actions, aim to minimize a variational free energy with respect to a probability distribution over states that they prefer to find themselves in~\citep{van2020deep,millidge2021whence, fountas2020deep}. This preference distribution entails a biased (conditional) prior on environmental states, which acts as an attractor for planning future behavior. Once this preference (or soft goal) is learnt the rest of the architecture machinery will drive the system (agent-world) to generate object-centric actions that minimize the discrepancy of the current and the desired state.

\paragraph{Preference network} We assume that the preference distribution is given by $\tilde{p}(\{\genstate_{T>t}^{(k)}\}_{k\in 1:K}|\{\v{\lambda}^{(k)}_t\}_{k\in 1:K})=\prod_k \mathcal{N}(\tilde{\v{\mu}}^{(k)}, {\tilde{\v{\sigma}}{^{(k)}}^2} )$, where $T$ is some time horizon and $t$ denotes the present. Note that this distribution is conditioned on the agent's internal beliefs, via the variational parameters $\lambda^{(k)}$~\cite{Greff2019a_Iodine,veerapaneni2020entity,van2023object}, which is the set of latent variational parameters (means and (log-)variances) for the k-th object slot that parameterize $q(s^{(k)})$.

We implement this mapping through a set-structured MLP network $\phi$ (architectural details in Appendix~\ref{sec:network-architectures}), which preserves the order of objects from input to output, and is invariant to the number of objects:
\begin{gather}
    \v{\nu}^{(k)} = \phi_\text{enc}(\v{\lambda}^{(k)}), \quad
    \v{c} = \frac{1}{K} \sum_k \v{W}_\text{ctx}\v{\nu}^{(k)} + \v{b}_\text{ctx} \\
    \tilde{\v{\mu}}^{(k)}, {\tilde{\v{\sigma}}^{(k)}} = \phi_\text{dec}\left(  \v{\nu}^{(k)}, \v{c} \right)
\end{gather}
In words, each $\v{\lambda}^{(k)}$ is projected into an embedding space by an encoder network. Object-wise embedding vectors $\{\v{\nu}^{(k)}\}$ are then linearly transformed and aggregated into a global context vector $\v{c}$, which is appended to each object embedding. Finally, the concatenated object+context embeddings are passed through a decoder network, to obtain object-wise preference statistics. Conceptually, since OBR determines its current goals by repeatedly applying the same local operation to each object representation (rather than a global operation on the full state of the environment), this may be interpreted as a form of (proto-)symbolic reasoning.  

Note that the preference network depends on the world model in an unsupervised learning scheme. The latent space geometry that the world model ends up learning is unknown a priori.  Therefore, the preference network cannot learn a mapping within this latent space (from current to desired states), before this latent space has been defined. Interestingly, multiple preference models can be trained "on top of" the same world model, allowing fast acquisition of novel tasks within the same environment. Practically, the one main difference to a common value network or critic, is that our preference network does not assign scores to states, but rather furnishes the agent with a desired state conditioned on its current context. This obviates the need to unroll (many) possible futures to evaluate which of these would be more desirable.

\paragraph{Control} Given a preference distribution $\tilde{p}(\genstate)$\footnote{Note that in this notation, we do not condition the preference on the current state belief as above, as the planning procedure described here can be applied to any preference distribution.}, the linearized dynamics model enables the agent to plan actions efficiently in closed form (without the need of rollouts). Specifically, the action plan  $\v{\pi} = \begin{bmatrix} \v{a}_{t+1}, \ \hdots, \ \v{a}_{t+T} \end{bmatrix}^{\text{T}}$ is computed by minimizing the path integral of the variational free energy~\citep{millidge2021whence} over some future time horizon (similarly to model predictive control)\footnote{Note that the notation in this section omits the object index $k$ for legibility, as actions can be planned independently between objects (a strength of our approach).}:

\begin{gather}
    \v{\pi}^* = \argmin_\v{\pi} \sum_{\tau=1}^{T}D_\text{KL}\left(q(\genstate_{t+\tau}|\v{\pi}) || \tilde{p}(\genstate)  \right) = (\v{U}^T \v{L} \v{U} + \lambda_a \v{I})^{-1} \v{U}^T \v{L} \v{e}
    \label{eq:planning}
\end{gather}
Importantly, this optimization can be performed in closed form, by computing the discrepancy between the preferred state and the sequence of states that will unfold if no action is taken, and projecting this discrepancy ($\v{e}$) onto the pseudo-inverse of a matrix $\v{U}$ that maps actions within the planning horizon to their (cumulative) effects over that time window. $\v{L}=\text{diag}(\tilde{\v{\sigma}}{^{(k)}}^{-2})$ is a diagonal matrix containing the precisions (inverse variances) of the preference distributions, and $\lambda_a$ controls the strength with which actions are regularized (shrunk) towards to a zero-mean prior. Further details are provided in Appendix~\ref{sec:planning-details}.

\subsection{Training}
The perception module is trained using pre-generated 4-frame videos with sparse actions, to minimize a bespoke ELBO loss that drives the model to reconstruct video frames accurately, while employing representations that are consistent with the dynamics model. Full details on the training procedure may be found in Appendix \ref{sec:training-procedure}. The preference network is trained separately, in a self-supervised fashion, to learn the state-preference mapping---OBR does not copy any behavior from a teacher agent. The model is shown example task episodes where objects start in a random configuration with random velocities, and are simulated for a few frames without any goal-directed interference. In the final two frames, the objects are moved (by an oracle agent) to their target locations (with a final velocity of 0). The perception module encodes each video frame in the latent space, and the preference network is trained to minimize the discrepancy (KL divergence) between its preference prediction for the ``seed frames", and the representation of the target frame.

\section{Results}
\label{sec:results}

We evaluated OBR qualitatively and quantitatively focusing on analyzing its ability to perform conditional behavioural reasoning using visual cues (Figure \ref{fig:results:intro}a). This is, to evaluate the capacity of the agent to learn proto-symbolic rules that follow 1st order logic. An exemplary conditional rule is ``If there is a Half-torus \underline{move} Boxes to the Top-Left and cones to the Bottom-Left and Half-torus to Middle-Right otherwise \underline{move} Boxes and Cones to the Right"\footnote{While we use words to describe the rule the agent learns the rule through unsupervised learning and there is no textual description of them}. Furthermore, we analyzed its generalization for the number of objects and the adaptation to changes in the environment. As extra result, although it is not the focus of this work, an analysis of  world model learning (i.e., segmentation and video prediction) can be found in the Appendix~\ref{sec:appendix:perceptualperformance}.

\begin{figure}[hbtp!]
	\centering
        \includegraphics[width=0.99\textwidth]{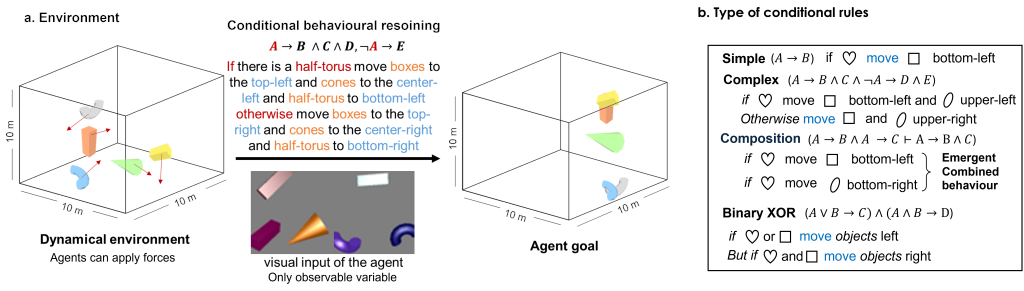}
        \caption{\textbf{Conditional behavioural reasoning experiments}. (a) 3D Active dSprites. The agent's visual input is the RGB image of the projected objects. Objects have 1st order dynamics with friction and the agent can apply forces. The behavioural reasoning is conditional on the objects present in the environment and the learnt rules. This is, the agent should move the objects differently depending on the proto-symbolic rules and the presence of objects. (b) Type of conditional rules that the agent should be able to learn: Simple and complex conditional rules, composition of two learnt rules and advanced logic XOR rules.} 	
 \label{fig:results:intro}
\end{figure}

Figure \ref{fig:results:intro}b details the type of rules that the agent can learn. Conditional rules, such as $(A \to B) \land (\neg A \to C)$: ``If there is a Half-torus move Boxes to the Top-left and cones to the Center-left, otherwise \underline{move} Boxes to the Top-right". It can also perform logical composition $(A \to B) \land (A \to C) \vdash A \to (B \land C)$ by learning two different rules separately but then they emerge combined during execution. Furthermore, it can also learn XOR operations, such as $(A \lor B) \to C) \land (A \land B \to D)$. Once the agent learns the rules, reasoning transforms into a behavioural response that changes the environment towards its internal preference (Fig. \ref{fig:results:intro}a). Figure \ref{fig:results:intro}c shows an OBR agent solving several conditional reasoning instances with different number of objects in the scene.

\subsection{Environment}
We developed, Active dSprites, which is an ``activated" version of the various multi-dSprites datasets that have been used in previous work on object-based visual inference (e.g.~\citep{Greff2019a_Iodine,locatello2020SlotAttention}). Not only does it include (continuous) dynamics, but these dynamics can be acted on by an agent (through continuous control actions that accelerate the objects). Thus, active dSprites is an interactive environment, rather than a dataset. 
We implemented two different scenarios in the active dSprites environment. The first, objects are 2.5-D shapes (e.g., squares, ellipses and hearts)---they have no depth dimension of their own, but can occlude each other within the depth dimension of the image. The second objects are 3D shapes (e.g., boxes, half-torus, cones) that can move in a $10\times10\times10$ m cubic space. These objects are much more complex to learn than the 2D version, as they have lighting and shadows, producing color gradients that need to be encoded. To randomize the environment, when an active dSprites instance is intialized, object shapes, positions, sizes and colors are all sampled uniformly at random. Initial velocities are drawn from a Normal distribution. Shape colors are sampled at discrete intervals spanning the full range of RGB-colors, while a background color is drawn from a set of evenly spaced grayscale values between black and white. Shapes are presented in random depth order. In the 3D environment the lighting sources are fixed. Code for the active dSprites environment can be found at \href{https://github.com/neuro-ai-robotics/OBR}{github.com/neuro-ai-robotics/OBR}.

\subsection{Problem complexity analysis through baselines comparison}
We evaluated the problem solving complexity of a basic conditional rule "If Heart then \underline{move} Boxes to the Right, otherwise \underline{move} Ellipses to the Left" comparing different oracle and baseline algorithms with our approach~\ref{fig:results:baseline}. We evaluated different algorithms with full observability (i.e., with access to object locations): Linear-Quadratic Regulator control~\cite{kalman1960contributions, kirk2004optimal}, (implemented in the Python 3 control systems library), Soft-Actor-Critic (SAC)~\cite{haarnoja2018soft} and Proximal Policy Optimization (PPO)~\cite{schulman2017proximal} (both used from the stable baselines 3 library~\cite{raffin2021stable}), and our OBR with perfect access to the true environmental state). And with partial observability (i.e., the agent only has access to the RGB image (pixels): SAC, PPO and OBR. For RL algorithms we gave dense rewards using the object distance between the current state and the desired state. For the OBR we only allow actions after frame 5. Here objects are 2.5 dimensional and during training the shape, color and positions are randomized and velocities are sampled from a Normal distribution with mean 0.

\begin{figure}[hbtp!]
	\centering
        \includegraphics[width=0.95\textwidth]{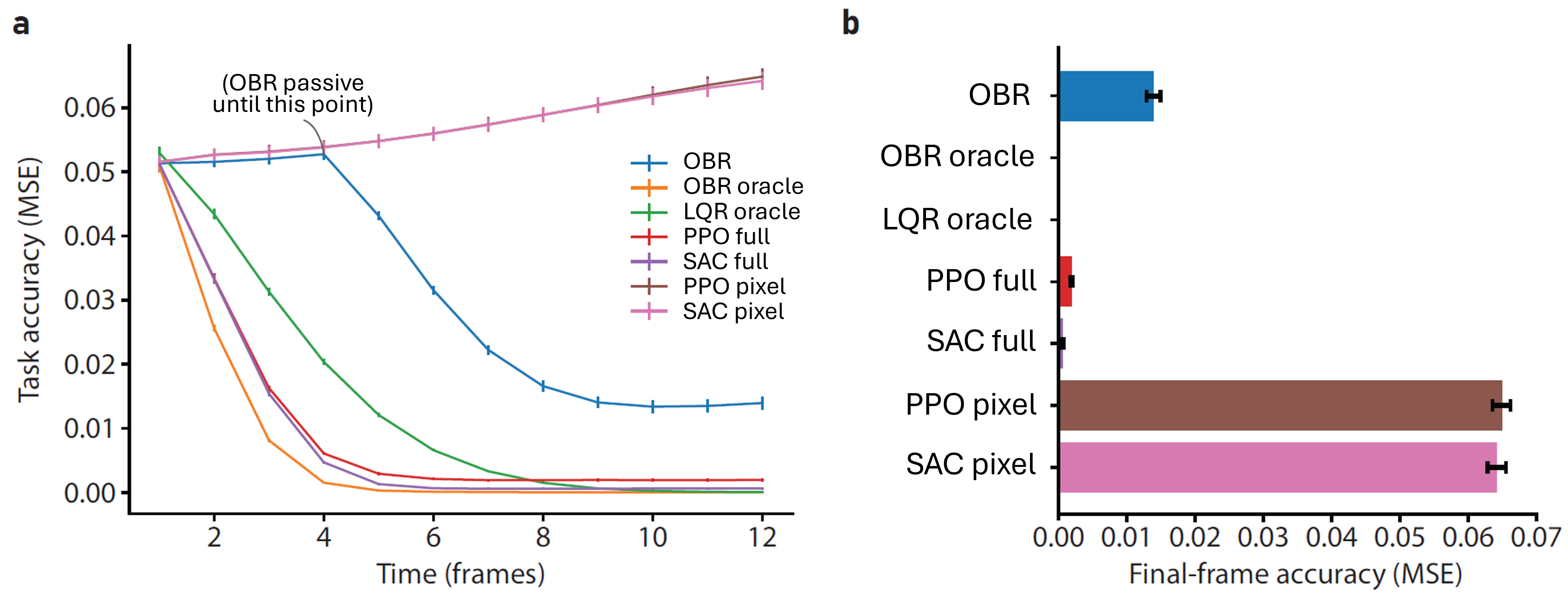}
        \caption{\textbf{Complexity analysis through baseline comparison}. Oracle algorithms have access to the objects location and dynamics, and define the upper-bound performance. RL algorithms (PPO and SAC) are tested with full observability and partial observability (RGB image pixels) (a) Algorithms performance at each frame of execution for a learnt single conditional rule $(A\to B) \land (\neg A\to C)$. Colored lines are the Mean Squared Error (MSE) between the objects location and the idealized goal location for 512 tested instances after training. Error bars are the standard deviation. (b) Average performance of the algorithms at the final frame. The lower the better.}
	\label{fig:results:baseline}
\end{figure}

Results are shown in Fig. \ref{fig:results:baseline}. The left panel shows the task accuracy (Mean Squarred Error, MSE) of the current object locations and the desired state in projected image)\footnote{Note that this error is over the ground truth goal as we can use the environment information to obtain the true image projection of the goal.}. All methods with full observability can solve the task, having all of them similar performance. When restricting the input to RGB images (pixels) standard SAC or PPO cannot learn the conditional behavioural rule. We can see that the MSE increases along the time. OBR, while it is not as accurate as the oracle, can learn the rule. The right panel shows the final performance of the compared algorithms. Oracle algorithms show the upperbound performance as it is constrained by the dynamics of the environment. 

These results first show, in line with previous works~\cite{veerapaneni2020entity}, that manipulating objects in the environment cannot be solved through non object-centric approaches, and highlights an important weakness of current pixel-based RL baselines. Second, it shows that a basic reasoning rule that implies interaction with the environment is already sufficiently complex. Thus, learning multiple rules and composition in this active dSprite synthetic environment is relevant. 

\subsection{Behavioural reasoning}

\subsubsection{Abstract behavioural reasoning with conditional rules}
\begin{figure}[hbtp!]
	\centering
        \includegraphics[width=0.95\textwidth]{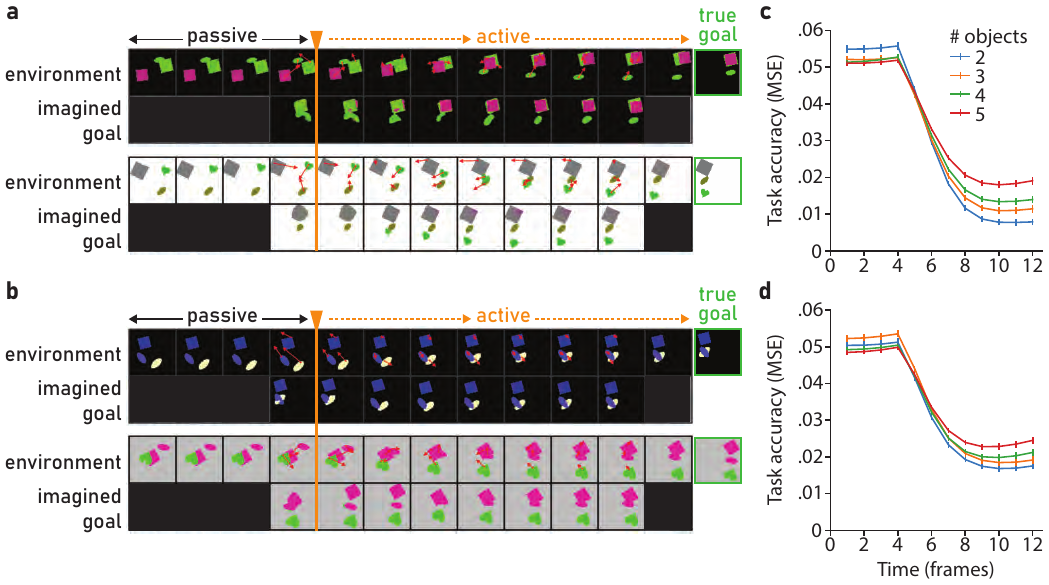}
        \caption{\textbf{Behavioural reasoning with 2D objects}. (a) Two instances of the active 2D dSprites solved by the OBR agent which has learnt the ``IfHeart" conditional rule: $($Heart $\to$ Squares to top-right $\land$ Ellipses to middle-right$) \land (\neg$ Heart $\to$ Squares to top-left $\land$ Ellipses to middle-left $\land$ Hearts to down-left$)$. Environment row shows the real scenario with the objects and Imagined goal row show the desired preference of the agent. The red arrows are the control actions applied to the objects in every frame. The agent is passive until the 5th frame and the preference network waits three frames to provide an output (this allows perception to be stable). The true goal is the idealized desired location of the objects. (b) Two instances of the active 2D dSprites solved by the OBR agent which has learnt the Heart XOR Square task. (c and d) Statistical evaluation of both tasks accuracy (MSE with respect to the idealized true goal) for different number of objects in the scene. The training was performed with 3 objects. Colored lines describe the MSE between ground-truth goals and actual object positions and velocities. Error bars reflect +/- 1 SEM.}
	\label{fig:results:nobjects}
\end{figure}

We evaluated the capacity of the architecture to learn proto-symbolic behavioural rules, and its generalization to different number of objects. We evaluated both 2D and 3D scenarios. Figure \ref{fig:results:nobjects} shows an OBR agent execution for an``IfHeart" conditional $(A \to B) \land (\neg A \to C)$ and the XOR rules, and two randomized scenarios for each rule. In Fig.~\ref{fig:results:nobjects}a, the first scenario, there is no heart, and boxes should be moved to the upper-left and ellipses to the middle-left. In the second one, there is a heart present, and boxes, ellipses and hearts should be moved to the top-left, center-left and bottom-left, respectively. Thus, an object's own shape determines its goal position along the vertical axis, while the presence of a heart anywhere in the scene determines all objects' goal positions along the horizontal axis. The agent is passive until the 5th frame. The environment row shows the visual input and the red arrows describe the forces applied. The imagined goal row, describes the output from the preference network at each frame. Note that the preference is in encoded only in the latent representation and here, for visualization purposes, we show its reconstruction using the decoder network. We let the agent to perform three frames of perceptual inference before activating the preference network to infer the current goal. The true goal is the idealized solution of this ``IfHeart" task. Analogously, in Fig.~\ref{fig:results:nobjects}b, the OBR agent, which has learnt the Heart XOR Square task, solves two 2d active dSprites scenarios.

Figure \ref{fig:results:nobjects}c,d shows the statistical performance of the OBR for different number of objects for the two conditional rules. The network is trained with three objects and then evaluated on two, three, four and five objects in the scene. Again here, the agent is passive until the fifth frame. The error curves shows that OBR interpret and manipulate the environment using the learnt conditional rule.

Figure \ref{fig:results:3d} shows an OBR similar analysis but in a 3D environment. The 3D scenario is considerably much more complex than the 2D one as objects have an extra dimension to move and colors have lighting. Here, the conditional rule learnt depends on the appearance of a HalfTorus object. The OBR preference network is also decoded to show its visual interpretation. Relevantly, the second instance shows that OBR can deal with the absence of multiples appearances of objects in the scene. There are two HalfTorus and no Cones in Fig.~\ref{fig:results:3d}a bottom instance. Figure \ref{fig:results:3d}b shows the statistical performance of the OBR for different number of objects. The network is again trained with three objects and then evaluated on two, three, four and five objects in the scene. The agent is passive until the fifth frame. The error curves (towards zero) shows that OBR interpret and manipulate the environment using the learnt conditional rule. Figure~\ref{fig:results:intro}c describes the 16 randomized OBR agent executions in 3D for 2,3,4 and 5 objects in the scence. For completeness, several randomized executions in 3D with different rules can be inspected at the github repository as animations.
 
\begin{figure}[hbtp!]
	\centering
        \includegraphics[width=0.99\textwidth]{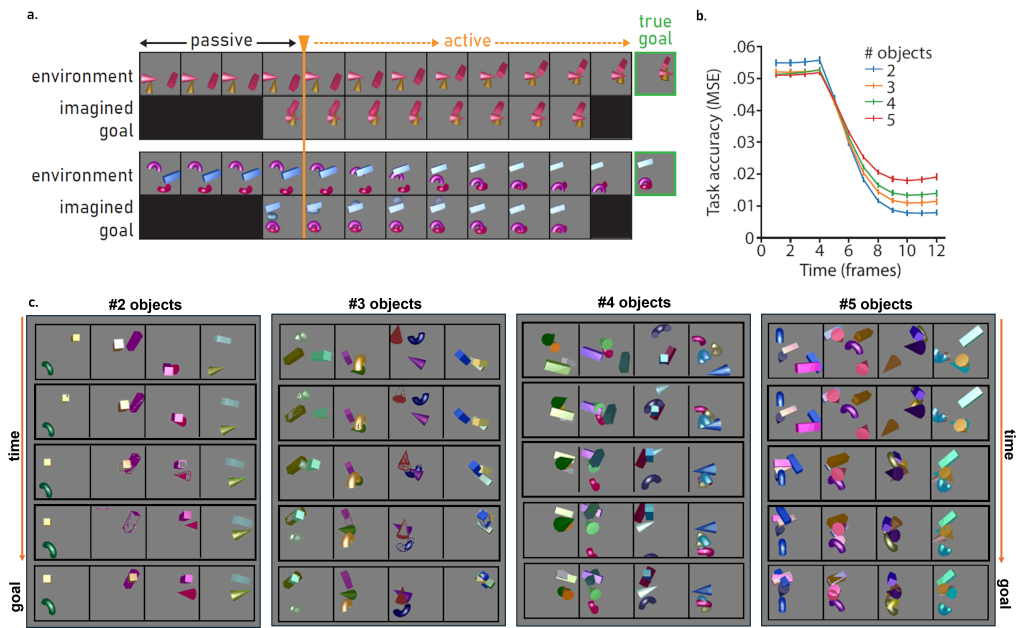}
        \caption{\textbf{Behavioural reasoning with 3D objects}. (a) Two instances of the active 3D dSprites solved by the OBR agent which has learnt the ``IfHalfTorus" conditional rule: $($HalfTorus $\to$ Boxes to top-right-middle $\land$ Cones to middle-right-middle$) \land (\neg$ HalfTorus $\to$ Boxes to top-left-front $\land$ Cones to middle-left-front $\land$ HalfTorus to down-left-front$)$. Note that in the second instance there are only Boxes and HalfTorus and \underline{no} Cones but the OBR solves the conditional task without troubles. Environment row shows the real scenario with the 3D objects and Imagined goal row show the desired preference of the agent. The red arrows are the control actions applied to the objects in every frame. The agent is passive until the 5th frame and the preference network waits three frames to provide an output (this allows perception to be stable). The true goal is the idealized desired location of the objects. (b) Statistical evaluation of the task accuracy (MSE with respect to the idealized true goal) for different number of objects in the scene. Blue lines describe the mean and the error bars denote the standard deviation. (c) Instances of the OBR agent sequential behaviour with an IfHalfTorus rule with different number of objects in the scene. The goal represents the ideal location of the objects given the rule. The desired goals of the agent (preferent internal state) are not shown and are learnt through unsupervised learning using the preference network.} 
	\label{fig:results:3d}
\end{figure}

\subsubsection{Adaptation to changes in the environment}
We evaluated the capacity of the OBR agent to recover from unexpected changes in the environment. For that purpose, we generate instances of the active dSprites that perform an object substitution in the middle of the execution. For instance, a Heart is changed by a Square during a conditional rule execution. Figure~\ref{fig:results:adapt}a describes two instances of the adaptation environment and Fig.~\ref{fig:results:adapt}b the statistical performance over 512 randomized instances. The object substitution experiment test the perception module, which needs to adapt fast to estimate the new situation, the preference network, which has to adapt the imagined goal depending on the objects that appear in the scene, and the control module that should provide the correct continuous actions to correct for the previously generated object movements. The true goal describes the idealized desired objects location after object substitution. The task error shows a decrease of performance until the system is able to recover and generate the proper behaviour to fulfill the conditional rule.

\begin{figure}[hbtp!]
	\centering
        \includegraphics[width=0.99\textwidth]{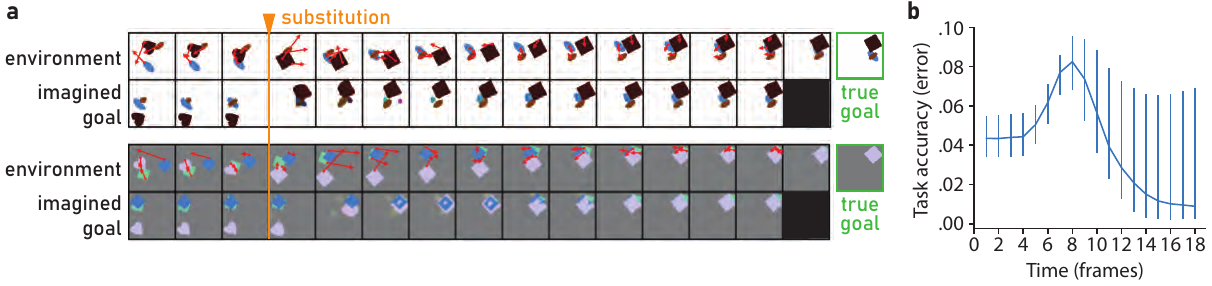}
        \caption{\textbf{Adaptation}. (a) Two instances of the active dSprites solved by the OBR agent which has learnt the ``IfHeart" conditional rule: $($Heart $\to$ Squares to top-right $\land$ Ellipses to middle-right$) \land (\neg$ Heart $\to$ Squares to top-left $\land$ Ellipses to middle-left $\land$ Hearts to down-left$)$. At 4th frame one of the objects (in this case the heart is substituted by a square, forcing not only to change the object-centric perception but the conditional behaviour, as the resulting reasoning depends on the appearance of the Heart. (b) Statistical performance evaluation of 512 instances of the object substitution experiment. The blue line is the median and error bars define the inter-quartile range to deal with the skewed data due to failures.}
	\label{fig:results:adapt}
\end{figure}

\subsubsection{Logical composition emergence}
We further evaluated the capacity of the agent to do logical composition $(A \to B) \land (A \to C) \vdash A \to (B \land C)$, as described in Fig.\ref{fig:results:compos}. We trained the preference reasoning module on two rules separately and then we evaluated in the testing phase the capacity of the agent to combine both. For instance, Rule 1, If Heart \underline{move} Squares to Top-left, otherwise, to Middle-right and Rule 2, If Heart \underline{move} Ellipses to Middle-left, otherwise, to Middle-right. In the testing phase, when there are both squares and ellipses the agent should compose both rules into: If Heart \underline{move} Squares to Top-left and Ellipses to Middle-left, otherwise to Middle-right. Fig~\ref{fig:results:compos} shows the task performance, computed as the MSE to idealized desired location of the objects, as if the two rules where being computed. The OBR agent is passive until the 5th frame. The plot shows that the performed actions on the objects actually solve the task, confirming the that agent reasoning combines both rules. 
\begin{figure}[hbtp!]
 	\centering
         \includegraphics[width=0.90\textwidth]{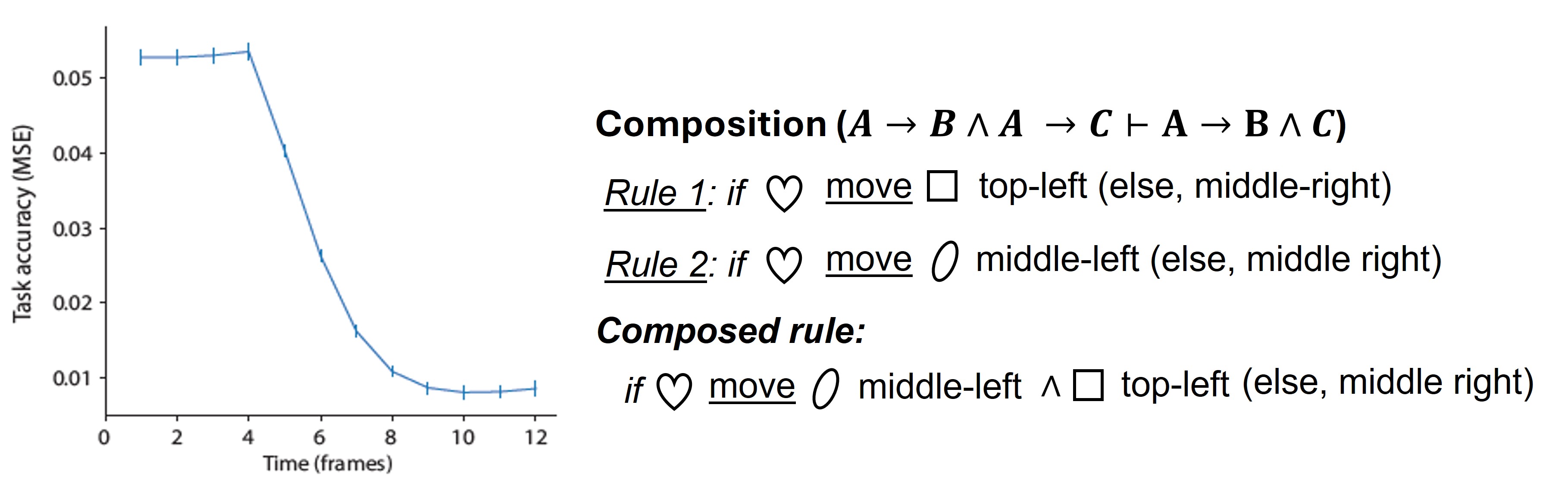}
    \caption{\textbf{Logical composition}. 
    The OBR agent's preference reasoning module is trained on two rules separately and evaluated at test time on scenes where both conditions apply. 
    For Rule~1, the agent learns to move Squares to the top-left when a Heart is present, and to the middle-right otherwise. 
    For Rule~2, it learns to move Ellipses to the middle-left when a Heart is present, and to the middle-right otherwise. 
    At test time, the agent must combine both mappings. The composed behavior, in which Squares move to the top-left and Ellipses to the middle-left (else, to the middle-right), is not provided during training and therefore emerges during inference. 
    Task accuracy over 512 instances, measured as the MSE with respect to the ideal target locations, confirms that the agent recovers the composed rule during execution.}
 	\label{fig:results:compos}
 \end{figure}

\subsection{Systematic Ablation Study and Comparison}

To understand how different components contribute to object-centric behavioral reasoning, we decompose OBR into three functional blocks: perception (object representation learning), dynamics (temporal prediction), and action (control generation). 
We evaluate each block in isolation to quantify its role and to clarify how specific architectural choices influence overall performance.

\subsubsection{Perceptual module}

A central design choice in object-centric perception is the form of inference used to obtain object representations. Iterative methods refine these representations through multiple optimization steps per frame, enabling adaptive correction of early estimates and the integration of information across time. Approaches such as the proposed OBR, OP3~\cite{veerapaneni2020entity}, and IODINE~\cite{Greff2019a_Iodine} belong to this family of iterative amortized inference methods. Fully amortized alternatives compute object representations in a single feedforward pass and prioritize computational efficiency. The dominant trend in recent literature extends this amortized direction, particularly through Slot Attention~\cite{locatello2020SlotAttention} and its variants.

We compare our iterative perceptual backbone, which extends the IODINE architecture for video prediction and is referred to as IODINE-dyn, to the fully amortized SAVi-dyn~\cite{kipfconditional_SAVi} across three dimensions: reconstruction quality, temporal prediction, and perceptual adaptation. 
Models were evaluated on 1000 test sequences from the Active DSprites 2D environment using three standard metrics: LPIPS for perceptual reconstruction quality, mIoU for pixel-level segmentation accuracy, and FG-ARI for clustering consistency of foreground object assignments (Table~\ref{tab:quantitative-perceptual}). 
Across these global metrics, IODINE-dyn achieves consistently higher performance, supporting the effectiveness of iterative inference under dynamic conditions.

Beyond aggregate metrics, the models differ in how their inference mechanisms respond to challenging visual situations.  In our dataset, several factors such as small object size, muted colors, weak contrast, and partial occlusions are common, making some objects difficult to extract from a single frame. 
Iterative inference enables IODINE-dyn to progressively refine uncertain hypotheses as more evidence becomes available, often resolving initially ambiguous shapes over successive frames.  Fully amortized inference, as used in SAVi, tends to produce stable representations but is less able to revise early commitments, which can lead to incomplete or imprecise segmentations when cues remain weak throughout a sequence. 
This pattern is reflected in the per-frame trends in Fig.~\ref{fig:challenge-segmentation}, where IODINE-dyn improves over rollout length while SAVi-dyn remains comparatively stationary.


\begin{table}[!htbp]
\caption{Quantitative results for slot dynamics prediction and unsupervised object discovery. 
Visual quality of predicted frames is measured by LPIPS (lower is better), while object discovery quality is measured by mIoU and FG-ARI (higher is better). 
Best results are in bold.}
\label{tab:quantitative-perceptual}
\renewcommand{\arraystretch}{1.2}

\resizebox{\textwidth}{!}{%
\begin{tabular}{@{}l
                p{3cm}
                p{3cm}
                p{3cm}
                p{3cm}
                @{}}
\toprule
\textbf{Method} &
\textbf{LPIPS} $\downarrow$ &
\textbf{mIoU} $\uparrow$ &
\textbf{FG-ARI} $\uparrow$ &
\\[-0.2em] 
\midrule

SAVi (8f) &
0.206 $\pm$ .014 &
0.309 $\pm$ .019 &
0.573 $\pm$ .041 &
\\

SAVi (16f) &
0.222 $\pm$ .014 &
0.576 $\pm$ .016 &
0.425 $\pm$ .034 &
\\

IODINE-dyn (ours) &
\textbf{0.054 $\pm$ .012} &
\textbf{0.632 $\pm$ .034} &
\textbf{0.758 $\pm$ .033} &
\\

\bottomrule
\end{tabular}
}
\end{table}

Quantitative results do not fully capture differences in shape fidelity. As shown in Fig.~\ref{fig:challenge-segmentation}, SAVi often produces representations that encode color and approximate location but do not preserve object morphology, which limits its ability to distinguish between objects with similar appearance. IODINE-dyn, in contrast, maintains sharper and more structured boundaries across frames. This distinction is functionally relevant: in conditional behavioral reasoning tasks, identifying the correct object shape is required to ground the appropriate rule, and reduced shape fidelity can therefore hinder downstream decision making.

\begin{figure}[!htbp]
    \centering
    \includegraphics[width=\linewidth]{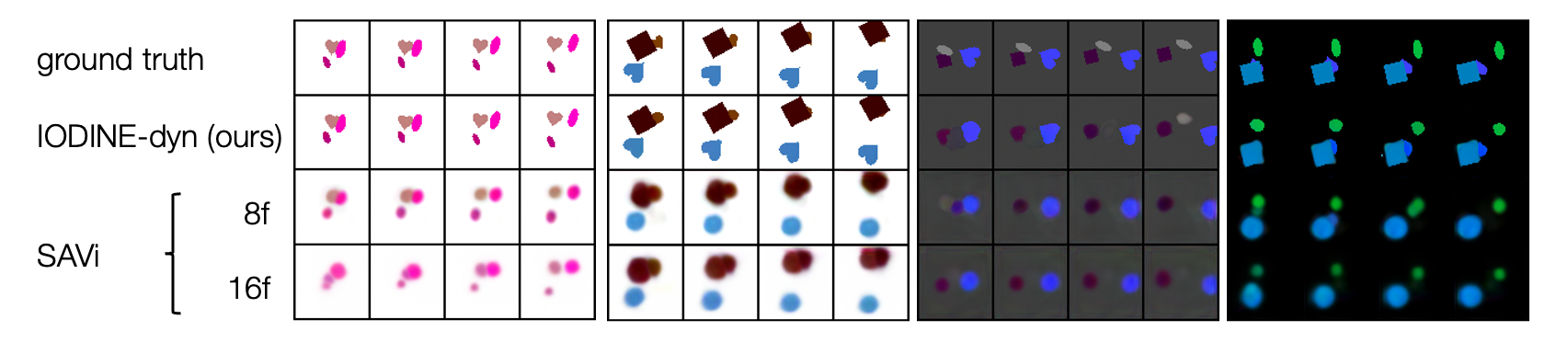}
\caption{\textbf{Challenging perceptual cases.}
We present a representative subset of sequences from the unsupervised object discovery task that are naturally difficult for object-centric models due to factors such as small object size, color similarity, low contrast, and partial occlusions.
Rows correspond to ground-truth frames, IODINE-dyn and SAVi reconstructions using 8-frame and 16-frame refinement, respectively.
In these conditions, SAVi often prioritizes color over geometry, leading to incorrect slot assignment, accidental merging or splitting of objects, or temporally stable yet inaccurate representations. Occlusions further amplify these errors, causing the model to miss partially hidden shapes or to misplace low-contrast objects, often producing blurred or inconsistent estimates.
IODINE-dyn (ours) performs iterative refinement at each timestep. Ambiguous shapes gradually emerge as evidence accumulates, and object identity is preserved more consistently across frames.
Across these examples, the figure illustrates how the models behave under challenging visual conditions and highlights the qualitative differences in their treatment of object geometry and occlusion.}

    \label{fig:challenge-segmentation}
\end{figure}

Because SAVi often fails to recover object morphology beyond coarse Gaussian-like blobs, a shape-based substitution analysis is not informative for this model class. To assess adaptation under conditions where both methods can maintain object identity, we instead performed a color-based intervention. At frame~2, one object was replaced by an identical shape with a different color, and we measured how accurately each model adapted to this change using a color fidelity score, defined as the normalized Euclidean distance in RGB space (0 to 1, lower is better). This setup parallels the object substitution experiment but isolates color as the causal factor. 

Across many sequences, IODINE-dyn adapts rapidly to the intervention, with error decreasing immediately after the change and continuing to improve over time. SAVi adapts more slowly. The 8-frame variant shows moderate improvement, while the 16-frame variant exhibits more variability across slots and often fails to consistently propagate the new appearance. We include both variants to illustrate the trade-off between temporal averaging and representational stability within fully amortized inference. Normalized RGB-distance values provide a quantitative reference for this pattern: immediately after the intervention (frame 1), the errors are 
0.067±0.081 for IODINE-dyn, 0.268±0.110 for SAVi-8f, and  0.443±0.135 for SAVi-16f; by frame 3 they decrease to 0.037±0.045, 0.202±0.071, and 0.299±0.188, respectively. Overall, these results show that iterative refinement provides a clear advantage in handling abrupt changes in appearance, complementing its benefits for recovering shape-based distinctions (Fig.~\ref{fig:color-intervention}).

 \begin{figure}[h]
    \centering
    \includegraphics[width=\linewidth]{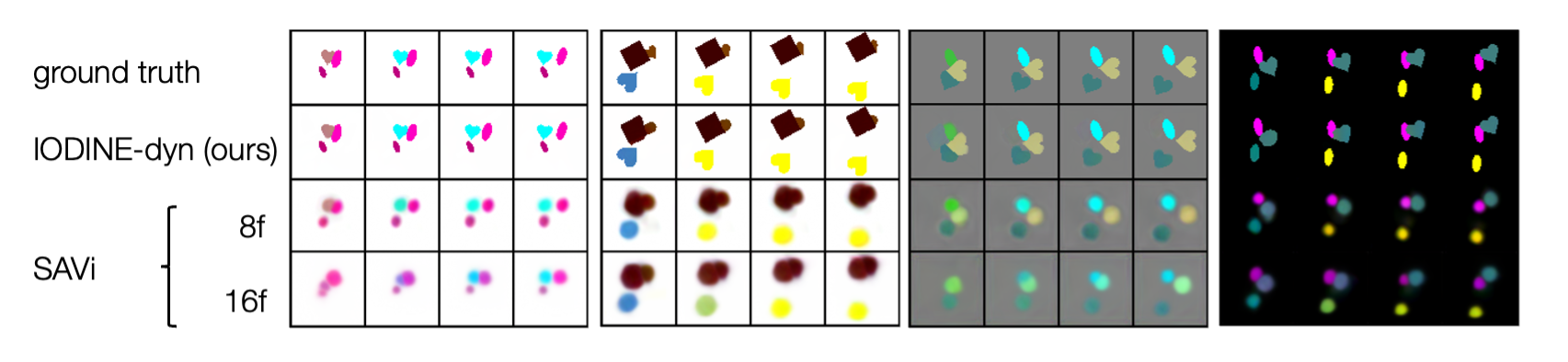}
\caption{\textbf{Color intervention.} To evaluate adaptation to sudden changes in appearance in the unsupervised object discovery setting, one object is replaced at frame~2 by an identical shape of a different color. The figure presents multiple sequences showing ground-truth frames, IODINE-dyn reconstructions, and the 8-frame and 16-frame variants of SAVi. IODINE-dyn (ours) adapts quickly and maintains stable representations, while SAVi adjusts more slowly and may introduce inconsistencies across slots when integrating the new color. Quantitative RGB-distance measurements follow the same pattern and support the qualitative differences observed here. }
    \label{fig:color-intervention}
\end{figure}

\subsubsection{Dynamics learning}

A central question in object-centric reasoning is how temporal dependencies are encoded. To disentangle perception from dynamics, we evaluate models using ground-truth object states as input. This upper-bound analysis isolates the contribution of the dynamics model itself and removes perceptual noise and control-feedback effects. 

We systematically compared four classes of dynamics architectures that span the main paradigms in structured object-centric learning: linear, recurrent, sparse-gated, and attention-based models. Linear approaches assume constant-coefficient state transitions and provide an interpretable baseline for temporal evolution. Recurrent architectures (GRU, LSTM) capture nonlinear dependencies either per object or jointly across the full system. Sparse-gated variants, inspired by event-driven mechanisms such as GateL0RD~\cite{gumbsch2021sparsely}, selectively activate updates to improve temporal credit assignment. Attention-based dynamics (transformer variants)~\cite{locatello2020SlotAttention} introduce relational reasoning across objects through learned cross-object interactions.

In the full OBR architecture, an action network infers object-centric control signals from pixel-space action fields, and a goal network generates desired states conditioned on context. The dynamics component itself uses a second-order linear state-space transition with Gaussian noise, a standard assumption in state estimation. The inverse mapping from actions to state updates is learned with a neural network. 

To evaluate these mechanisms, we organized the analysis into two experimental groups. Group A (Fig.~\ref{fig:group-a}) examines dynamics quality through multi-step prediction, permutation equivariance, and robustness to action noise. Group B (Fig.~\ref{fig:group-b}) assesses reasoning and generalization through counterfactual, interventional, and opposite-rule tests. Together, these experiments characterize how each architecture captures temporal structure, relational dependencies, and causal disentanglement in the object-centric state space.

Across permutation equivariance, noise robustness, and temporal extrapolation, the OBR linear state-space model shows consistently strong behaviour, delivering competitive performance across all tests while maintaining near-zero equivariance error (Fig.~\ref{fig:group-a}). In contrast, recurrent baselines show the anticipated tendency to accumulate prediction error over longer horizons, and attention-based models demonstrate no consistent benefit despite higher complexity. Overall, OBR dynamics delivers reliable accuracy with substantially fewer parameters, making it an appealing choice when latency and stability matter.

\begin{figure}[h]
  \centering
  \includegraphics[width=0.95\linewidth]{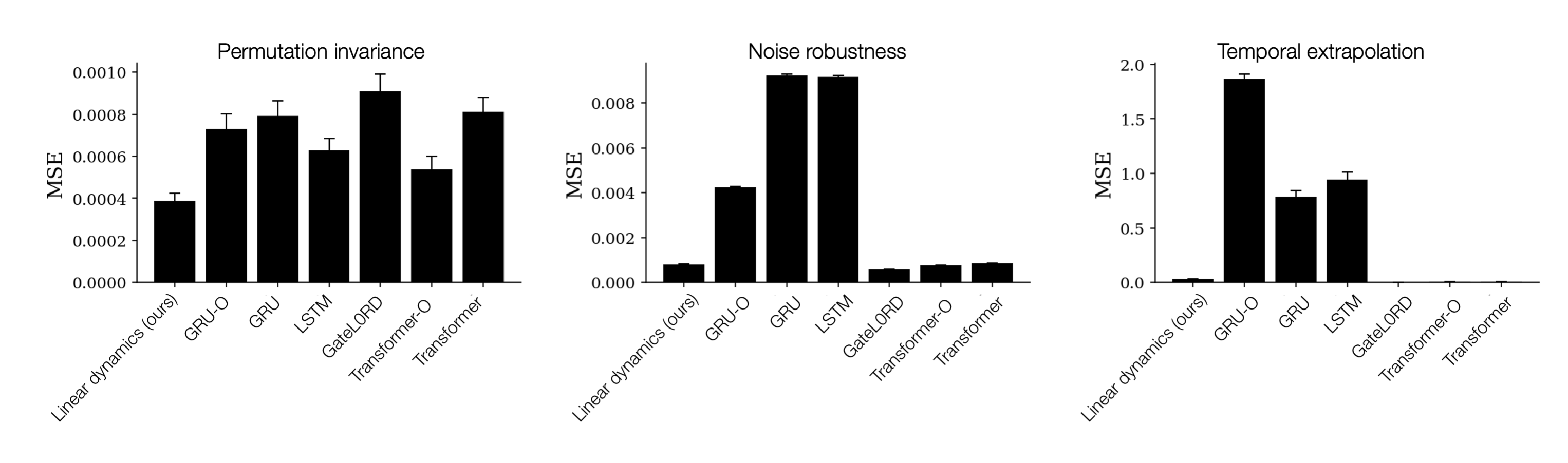}
  \caption{\textbf{Dynamics quality}. All dynamics models were trained on 8,000 trajectories of 50 steps each (400k state–action transitions) generated under the environment's Newtonian physics. Episodes begin from randomized initial conditions (positions uniformly sampled, velocities drawn from a zero-mean Gaussian, and fixed object identities), while actions correspond to uniformly sampled accelerations that drive deterministic kinematic updates. This training setup provides broad state–action coverage without policy bias and isolates the quality of the learned transition models. The figure reports mean squared error for three evaluations: permutation equivariance, noise robustness, and temporal extrapolation. The permutation test assesses whether predictions remain consistent when the order of objects in the state representation is permuted and then restored. The noise test measures prediction stability under small perturbations to the control inputs during short teacher-forced rollouts. The extrapolation test evaluates multi-step autoregressive prediction accuracy over longer horizons. Across these diagnostics, the OBR linear state-space dynamics achieve near-perfect permutation equivariance and competitive performance on noise robustness and extrapolation, while several recurrent models show greater sensitivity to object ordering and error accumulation at longer horizons.}
  \label{fig:group-a}
\end{figure}

Under online adaptation during 10-step rollouts (one gradient update per prediction), the OBR dynamics model attains among the lowest errors across three out-of-distribution tests that probe whether the models learned the underlying Newtonian physics rather than memorized training correlations. The opposite-rule test swaps object shape labels while leaving physical states unchanged; since shape does not influence motion, low error indicates that the model learned pure kinematics independent of task-specific labels. The counterfactual test exchanges two objects’ initial positions and velocities, evaluating whether the model can reason over alternative initial conditions. The interventional test freezes one object’s state (fixed position and zero velocity), probing whether the model respects the independence of object dynamics. Heavier recurrent and attention-based architectures often require careful tuning for online stability and show more variable behavior, whereas the linear state-space dynamics provide a reliable and lightweight option for adaptive closed-loop settings.

\begin{figure}[h]
  \centering
  \includegraphics[width=0.95\linewidth]{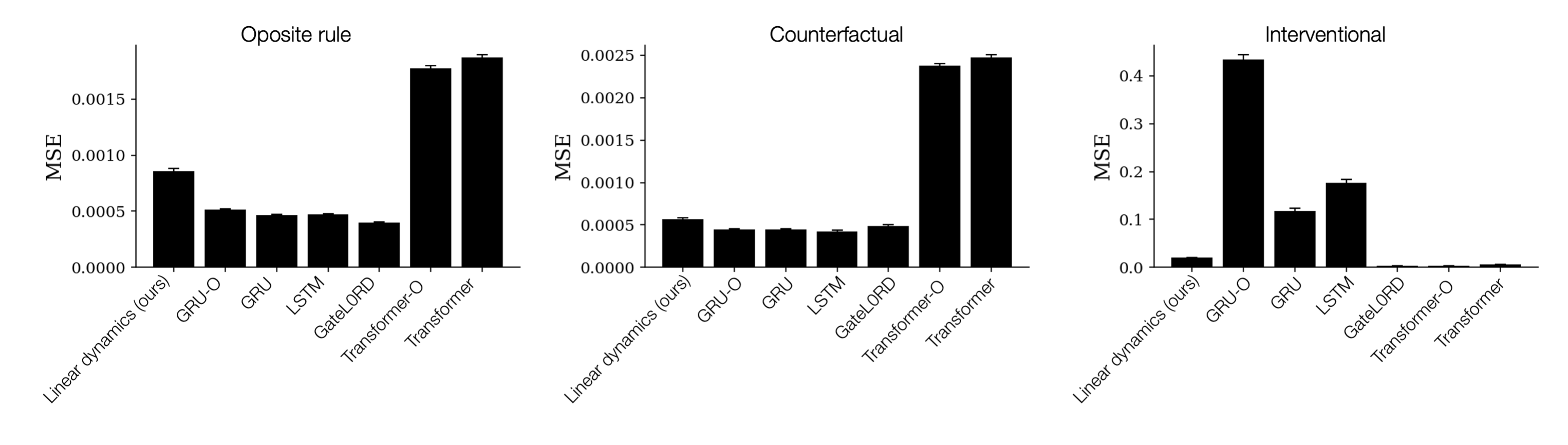}
  \caption{\textbf{Reasoning and generalization with online adaptation.} All dynamics models were trained under identical conditions on 8,000 trajectories of 50 steps generated from the environment's Newtonian physics. During evaluation, models perform one gradient update per timestep to adapt their predictions online. The figure reports mean squared error for three out-of-distribution tests that probe whether models learned the underlying physical dynamics rather than memorized training correlations. 
The opposite-rule test swaps object shape labels while keeping physical states unchanged; the counterfactual test exchanges the initial states of two objects; and the interventional test freezes one object's motion. Across these evaluations, the OBR linear dynamics model maintains competitive accuracy, while larger attention-based models often require careful tuning and exhibit more variable behaviour.}
  \label{fig:group-b}
\end{figure}

\subsubsection{Action module}
Finally, we compared OBR's action module and closed-form control (inspired by active inference-based planning ~\cite{lanillos2021active}) against the current monolithic state-of-the-art for visual world models with control, V-JEPA 2-AC ~\cite{assran2025v}. Both algorithms were evaluated on the conditional rule $(\text{A} \to \text{B}) \land (\neg \text{A} \to \text{C})$, referred to as the $\textit{If-Heart}$ task.

To avoid bias from implementation-specific training procedures in the native V-JEPA loss formulation, we assess three actor-head variants of V-JEPA 2-AC: the base model, a Dataset Aggregation (DAgger) variant, and a behavioral cloning (BC) variant. All V-JEPA variants share the same frozen encoder–predictor backbone and were trained under identical baseline conditions (complete architectural and training configurations are reported in Appendix~\ref{sec:training-procedure}). 

Figure $\ref{fig:sota-comparison}$ reports the evolution of prediction error over rollout length. OBR oracle, trained with oracle demonstrations (i.e., oracle-provided accelerations, identical to the supervision used by V-JEPA), achieves near-perfect performance (MSE $\approx$ 0.00001). The behavioral cloning variant of V-JEPA (BC) reaches MSE $\approx$ 0.021 under identical training conditions. Standard OBR achieves MSE $\approx$ 0.013 using self-generated demonstrations from its action and preference modules. The performance difference reflects architectural distinctions. OBR employs object-based representations and modular components (action and preference modules) that generalize across different conditional rules with the same trained mechanism. In contrast, the V-JEPA BC variant exhibits convergence in this rule, but requires oracle-provided actions (accelerations) during training. More broadly, V-JEPA operates on monolithic pixel-level representations and requires oracle demonstrations for each new rule.\\
 \begin{figure}[hbtp!]
    \centering
\includegraphics[width=0.99\linewidth]{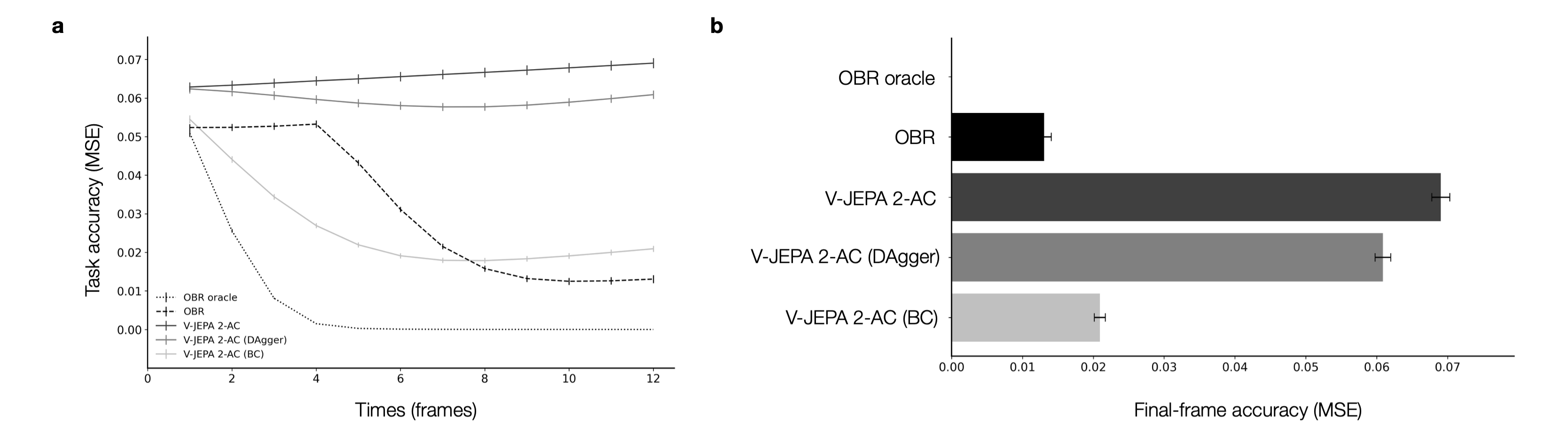}
    \caption{\textbf{State-of-the-art comparison on behavioural reasoning.}
    We evaluate OBR and three actor-head variants of the action-conditioned visual world model V-JEPA 2-AC (base, DAgger, BC) on the conditional rule (A$\to$B) $\land$ ($\neg$A$\to$C), referred to as the \textit{If-Heart} task. We include OBR oracle, trained with oracle demonstrations (identical training setup to V-JEPA variants). The metric is the Mean Squared Error (MSE) between the object's predicted location and the idealized goal location. Panel (a) reports error over rollout length for 512 test instances, and panel (b) shows the final-frame error. Lower is better. OBR oracle establishes the performance upper bound when trained with oracle demonstrations. Standard OBR, trained with self-generated demonstrations, achieves lower error than all V-JEPA variants.}
    \label{fig:sota-comparison}
\end{figure}

\section{Discussion}
\label{sec:conclusion}

We described a brain-inspired deep learning architecture that allows grounded conditional abstract reasoning and behaviour by leveraging object-centric learned representations as the units of ``mental" manipulation. Both perceptual and proto-symbolic goals were learnt in an unsupervised fashion from visual information (pixels). Results show that the system can learn and resolve conditional behaviours similar to 1st order logic: simple $A\rightarrow B$, complex $(A\rightarrow B \wedge C) \wedge (\neg A\rightarrow D \wedge E)$, XOR operation $(A\lor B\rightarrow C), A\wedge B\rightarrow D)$ and logical composition $(A\rightarrow B, A \rightarrow C \vdash A\rightarrow(B \wedge C)$. We analysed the model generalization to different number of objects (both in 2D and 3D), objects properties (e.g., color, shapes) and, critically, adaptation to changes in the environment (e.g., object substitution) thanks to the iterative inference and the preference network. 
OBR shows a promising direction in objects-based behavioral reasoning from pixels, mainly placing the foundations of grounded object-centric representations as a key inductive bias during learning, perception, reasoning and control. 

\subsection{Object-centric representations as symbols}
We showed that connectionist representation learning with inductive-bias, in particular object-centric representations, can work as a bridge between symbols and behaviour. Because objects are entities that live in both the real world and the internal representation of the agent, we can learn trajectories in the belief space that produce first-order-logic-like behaviours, where variables are substituted by object representations. Logical operators are encoded by learning the latent desired states (or preferences) of the agent and how to traverse the latent manifold to go from the current state to the desired state using the generative model.

\subsection{Grounded protosymbols and preferences in the latent space}
A key feature of OBR is that it learns grounded representations and manipulates them in the latent manifold, as well as generating the agent preferences (goals) in the object-centric latent representation. Agents' internal goals are tied to the world and the actions that it can perform in it. Hence, bringing benefits but also restrictions as it forces the system to perform embodied reasoning.

Learned protosymbolic rules depend on the potential learnable object features and how the agent can act on them---See latent space traversals in Appendix~\ref{sec:appendix:traversals}. For instance, the agent cannot learn to change the color of an object if all objects are in grey-scale or there is not a policy that can cause the desired effect. As a consequence, conditional behaviours that cannot be interpreted in the objects features space cannot be properly resolved in execution. On the good side, goal-conditioned behaviours are always grounded and therefore the gap between reasoning and control vanishes.

\subsection{Shortcomings and future work}
Following the brain inspiration, we employed an iterative inference procedure to endorse adaptation. Modules ablation results show that OBR's perception backbone (IODINE-dyn) displays better morphology discrimination and adaptation to changes in shape and colors than other state-of-the-art slot attention models. Anyhow, object-centric architectures are still too constrained to the discrimination accuracy of the perception module. While variational iterative inference offers more powerful per-scene reasoning and can refine ambiguous decompositions, amortized approaches may offer faster inference at the cost of reduced shape fidelity, suggesting that future work may benefit from hybrid strategies that balance accuracy and efficiency. Large models could help with the scaling to naturalistic images, but objects manipulation in the latent representation (e.g., JEPA or our OBR approaches) is critical to provide the needed abstraction against noise in the perceptual segmentation. 
Second, dynamics learning in OBR assumes that actions in the latent representation are linear plus noise. While this could fit a hierarchical interpretation of cognition, it may prevent complex non-linear manipulation dynamics. Ablation results shows that the dynamics learning in OBR can be improved. While OBR can recover from collisions, it cannot plan ahead with them.
Third, the achieved agent behaviours are much less complex than of current SOTA LLMs/VLM approaches, such as RT architectures~\cite{gu2023rt}. OBR provides, however, goal-conditioned behavioural reasoning, which is a major challenge in the field~\cite{sundaresan2024rt}. Our comparison reveals that monolithic architectures like V-JEPA 2-AC ~\cite{assran2025v} struggle with conditional rules even when trained with oracle demonstrations, while OBR's modular design achieves better performance using only self-generated demonstrations. This difference stems from architectural properties. OBR's preference network proposes context-based desired latent goals, enabling compositional generalization across rules with the same trained mechanism. In contrast, monolithic representations lack interpretable structure in the latent space, limiting their capacity to flexibly recombine learned components for novel behavioral requirements. In any case, increased expressivity of the reasoning should be investigated, for instance, by using OBR as the interface between symbolic (e.g., graph-based, language) and subsymbolic representations. Finally, and very relevant to operationalize this approach for robotics applications, the agent needs to include the body restrictions of performing an action in the world~\cite{lanillos2021neuroscience,taniguchi2023world}. This can be done by changing the action field mapping by a robotic arm controller.

\subsubsection*{Acknowledgments}

This work has been funded by the SPIKEFERENCE project, co-funded by the Human Brain Project (HBP) Specific Grant Agreement 3 (ID: 945539). AL was funded by the DeepSelf project, Deutsche Forschungsgemeinschaft (DFG), NUM: 467045002.

\section{Appendix}
\label{sec:appendix}

\subsection{Network architecture}\label{sec:network-architectures}
OBR's perceptual inference network consists of two separate Iterative Amortized Inference (IAI) modules, each of which in turn contains a refinement and a decoder module. The first IAI module concerns the inference of the state beliefs $q(\{\v{s}^{\dagger(k)}\})$ -- we term this the \textit{perceptual inference module}. The second IAI module infers the object action beliefs $q(\{\v{a}^{(k)}\})$, and we refer to this as the \textit{action inference module}. In addition, OBR comprises a preference network that maps the current latent state beliefs to a predicted preference distribution over future object states. Code for the OBR architecture can be found at \href{https://github.com/neuro-ai-robotics/OBR}{github.com/neuro-ai-robotics/OBR}.

\subsubsection{Perceptual inference module}
This module used a latent dimension of 16. Note that, in the output of the refinement network, this number is doubled once as each latent belief is encoded by a mean and variance, and then doubled again as we represent (and infer) both the states and their first-order derivatives. In the decoder, the latent dimension is doubled only once, as the state derivatives do not enter into the reconstruction of a video frame. As in \citep{Greff2019a_Iodine}, we use a spatial broadcast decoder, meaning that the latent beliefs are copied along a spatial grid with the same dimensions as a video frame, and each latent vector is concatenated with the $(x,y)$ coordinate of its grid location, before passing through a stack of transposed convolution layers. Decoder and refinement network architectures are summarized in the tables below. The refinement network takes in 16 image-sized inputs, which are identical to those used in \citep{Greff2019a_Iodine}, except that we omit the leave-one-out likelihoods. Vector-sized inputs join the network after the convolutional stage (which processes only the image-sized inputs), and consist of the variational parameters and (stochastic estimates of) their gradients. 

\vspace{12pt}
\noindent\normalsize{\textbf{Decoder (states)}} 
\\*\\*
\noindent\begin{tabular}{lccl}
  \toprule 
  \bfseries Type & \bfseries Size/\#Chan.~~ & \bfseries Act. func.~~ & \bfseries Comment \\
  \midrule 
  Input ($\v{\lambda}$) & 32 &  & \\ 
  Broadcast & 34 & & Appends coordinate channels \\
  Conv$^T$ $5 \times 5$ & 32 & ELU & \\
  Conv$^T$ $5 \times 5$ & 32 & ELU & \\
  Conv$^T$ $5 \times 5$ & 32 & ELU & \\
  Conv$^T$ $5 \times 5$ & 32 & ELU & \\
  Conv$^T$ $5 \times 5$ & 4 & Sigmoid or Softmax & Outputs RGB + mask\\
  \bottomrule 
  \addlinespace[6pt]
\end{tabular}

\vspace{12pt}
\noindent\normalsize{\textbf{Refinement network (states)}} 
\\*\\*
\noindent\begin{tabular}{lccl}
  \toprule 
  \bfseries Type & \bfseries Size/\#Chan.~~ & \bfseries Act. func.~~ & \bfseries Comment \\
  \midrule 
  Linear & 64 & & \\ 
  LSTM & 128 & tanh & \\
  Concat $[..., \v{\lambda}, \nabla_\v{\lambda}\mathcal{L}]$ & 256 & & Appends vector-sized inputs \\
  Linear & 128 & ELU & \\
  Flatten & 800 & & \\
  Conv $5 \times 5$ & 32 & ELU & \\
  Conv $5 \times 5$ & 32 & ELU & \\
  Conv $5 \times 5$ & 32 & ELU & \\
  Inputs & 16 & & \\
  \bottomrule 
  \addlinespace[6pt]
\end{tabular}

\label{sec:appendix:perceptualperformance}

\paragraph{Segmentation and prediction}
Figure~\ref{fig:segmentation} describes the internal mechanism of the perceptual module with the attentive masks segmentation and the objects prediction in the environment.
\begin{figure}[hbtp!]
	\centering
        \includegraphics[width=0.8\textwidth]{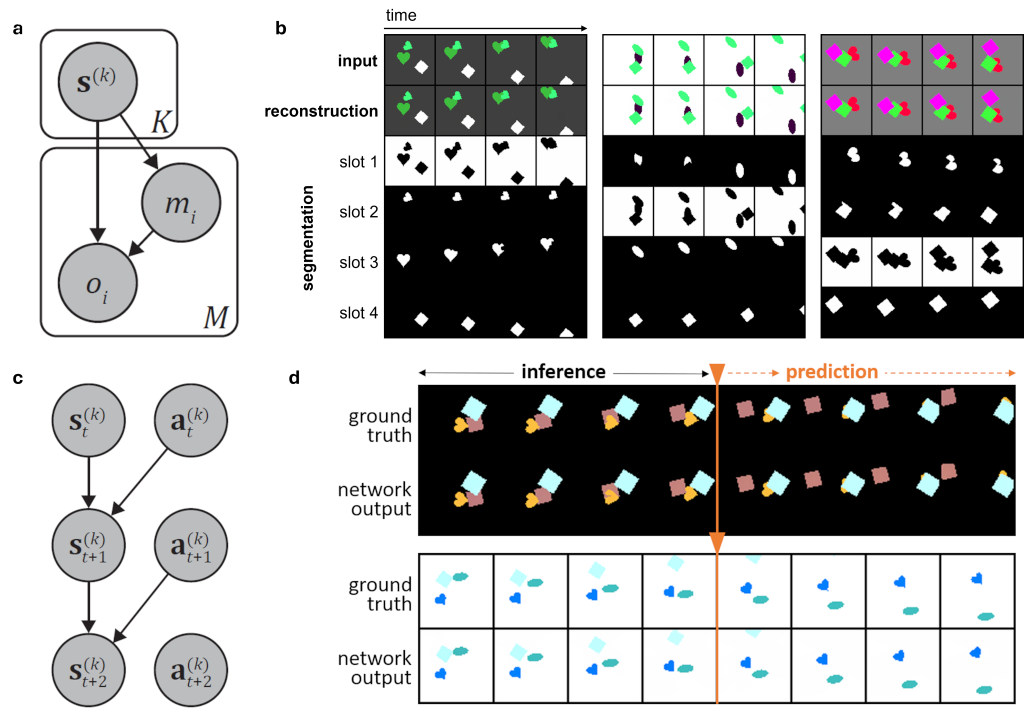}
        \caption{\textbf{Object-centric segmentation and prediction.} (a) Observation model (b) Three examples of attentive masks output by the OBR perceptual module with an IAI backbone. (c) Dynamics model. (d) Two examples of the perception predicting ahead in time the dynamics of the environment.} 
	\label{fig:segmentation}
\end{figure}

\paragraph{Prediction robustness}
To enable action planning across episodes of non-trivial length, it is important that OBR's perceptual inference and predictive abilities remain stable across longer time windows. In addition, a great benefit of OBR's slot-based architecture is that object slots can be added or removed at will, without having to learn additional connection weights~\cite{van2023object}. However, it is not a given that performance will be robust to such variations. Here, we test the robustness of OBR's world model, as measured by segmentation and reconstruction accuracy, to both of these dimensions (Fig.~\ref{fig:robustness}). Having been trained on 4-frame videos with 3 objects, OBR generalizes well to longer videos with fewer or more objects present in the scene. A slight drop-off in performance towards later frames in the 12-frame testing videos is easily remedied by a very short bout (3 epochs) of additional training with an additional set of longer (also 12-frame) training videos. Generalization to different numbers of objects is characterized by a slight drop in performance as the number increases, but this might also be (partly) attributed to increased complexity of the resulting images. 

\begin{figure}[hbtp!]
	\centering
        \includegraphics[width=0.8\textwidth, height=300px]{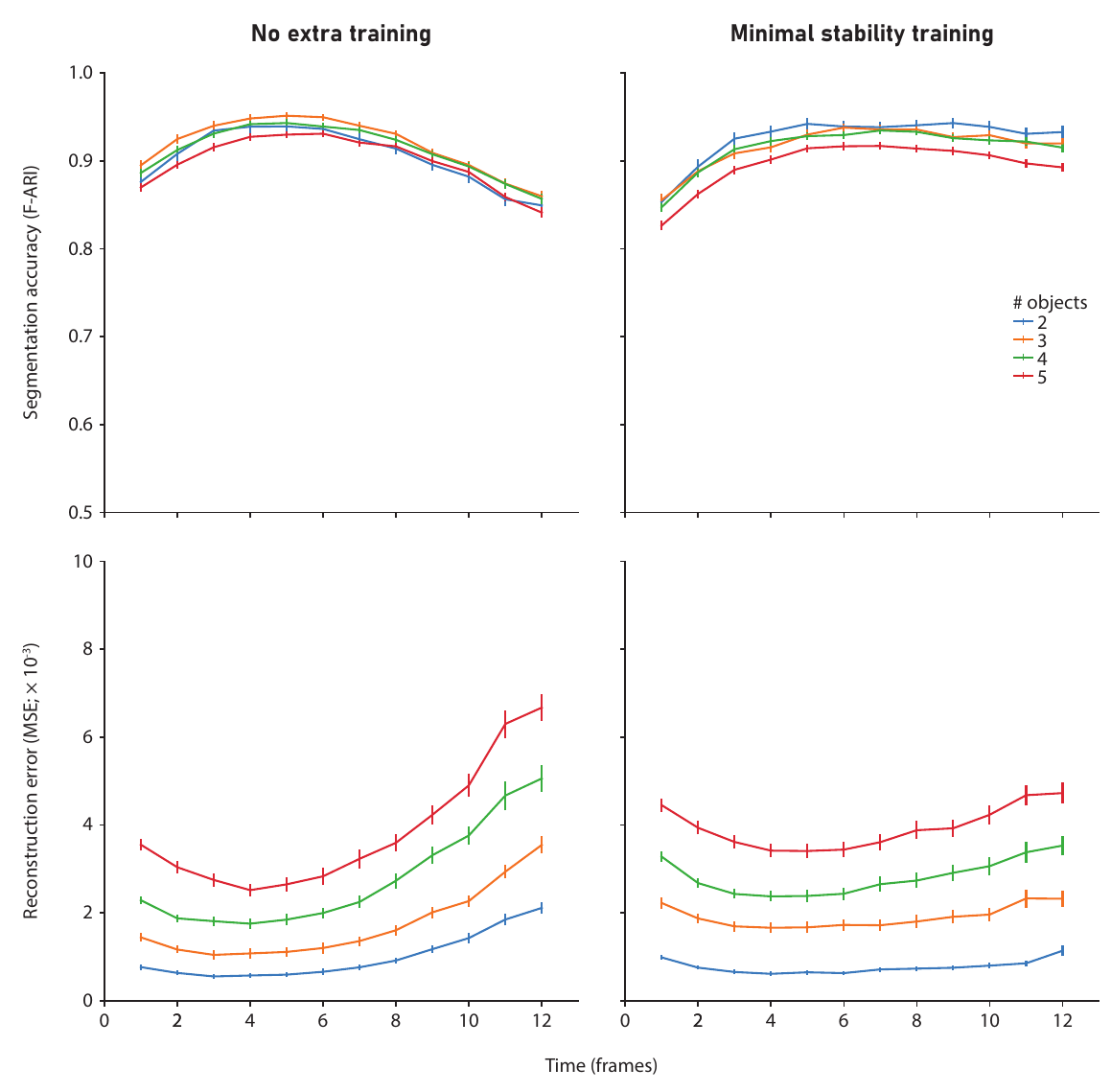}
        \caption{\textbf{Robustness to variations in video length and number of objects}. Segmentation accuracy (F-ARI score) and reconstruction error (MSE) across video frames that extend beyond the duration of the training videos (4 frames), for scenes with various numbers of objects (training videos contained 3 objects). Left panels show the performance of the network after training on 4-frame videos only. Right panels show the performance after just 3 epochs of additional training with 12-frame videos. Error bars depict the mean +/- 1 SEM.} 
	\label{fig:robustness}
\end{figure}

\subsubsection{Action inference module}\label{sec:action-inference-module}
A key challenge in multi-object environments such as the evaluated environment, is that an agent's internal representation of objects has an unknown (and possibly imperfect) correspondence to the objects in the environment. Even if the object features are inferred with perfect accuracy, the order of the objects in the representation is arbitrary. To solve this correspondence problem, we define an action space in which accelerations can be placed at pixel locations within the environment's image grid, and objects receive the sum of all accelerations that coincide with their visible pixels. Specifically, we introduce the notion of an \textit{action field} $\v{\Psi}=[\v{\psi}_1, ..., \v{\psi}_M]^T$: an $[M \times 2]$ matrix (with $M$ the number of pixels in an image or video frame), such that the $i$-th row in this matrix ($\v{\psi}_i$) specifies the (x,y,z)-acceleration applied at pixel $i$. The action on the $k$-th object is then given by:
\begin{equation} \label{eq:actionfielda}
    \v{a}_t^{(k)} = \sum_i [\mask_i=k]\v{\psi}_i
\end{equation}
where $\mask_i$ is a categorical variable that indicates which object pixel $i$ belongs to
\footnote{Note the use of Iverson-bracket notation; the bracket term is binary and evaluates to 1 iff the expression inside the brackets is true.}. This definition of actions in pixel space provides an unambiguous interface for the agent to interact with its environment.

The action inference module does not incorporate a decoder network, as the quality of the action beliefs is computed by evaluating equation \ref{eq:actionfielda} and plugging this into the ELBO loss from equation \ref{eq:elbo}. While this requires some additional sampling operations (see Appendix~\ref{sec:samplingtrick}), no neural network is required for this. This module does include a (shallow) refinement network, which is summarized in the table below. This network takes as input the current variational parameters $\v{\lambda_a}^{(k)}$ (2 means and 2 variances), their gradients, and the `expected object action', $\sum_i \hat{\mask}_{ik} \v{\psi}_i$. 

\vspace{12pt}
\noindent\normalsize{\textbf{Refinement network (actions)}} 
\\*\\*
\noindent\begin{tabular}{lccl}
  \toprule 
  \bfseries Type~~ & \bfseries Size/\#Chan.~~ & \bfseries Act. func.~~ & \bfseries Comment \\
  \midrule 
  Linear & 4 & & \\ 
  LSTM & 32 & tanh & \\
  Inputs & 10 & & \\
  \bottomrule 
  \addlinespace[6pt]
\end{tabular}\

\subsubsection{Preference network}

The preference network consists of encoder and decoder networks that operate on single object representations, as well as a context module that aggregates object embeddings produced by the encoder into a single context vector, which is then broadcast and appended to the object embeddings that are fed into the decoder. Details for the two-layer encoder and decoder networks are listed below. The context module consists of a single Linear layer with input and output sizes both equal to 64, and no activation function.

\vspace{12pt}
\noindent\normalsize{\textbf{Encoder}} 
\\*\\*
\noindent\begin{tabular}{lccl}
  \toprule 
  \bfseries Type & \bfseries Size/\#Chan.~~ & \bfseries Act. func.~~ & \bfseries Comment \\
  \midrule 
  Linear & 64 & ELU & \\
  Linear & 64 & ELU & \\
  Inputs & 64 & & \\
  \bottomrule 
  \addlinespace[6pt]
\end{tabular}

\vspace{12pt}
\noindent\normalsize{\textbf{Decoder (preferences)}} 
\\*\\*
\noindent\begin{tabular}{lccl}
  \toprule 
  \bfseries Type & \bfseries Size/\#Chan.~~ & \bfseries Act. func.~~ & \bfseries Comment \\
  \midrule 
  Linear & 64 & ELU & \\
  Linear & 64 & ELU & \\
  Inputs & 128 & & \\
  \bottomrule 
\end{tabular}

\subsection{Inference procedure}\label{sec:inference-procedure}
Inference is performed in a recurrent algorithm that loops through the decoder and refinement subnetworks. For a single video frame, the inference starts by initializing the object-wise latent beliefs. For the first frame in an episode, beliefs are initialized to a learned default vector of variational parameters $\v{\lambda}_0$. For subsequent frames, beliefs are initialized by extrapolating the inferred dynamics from the previous frame.

From each belief $q(\genstate^{(k)})$, we then (as in \citep{Greff2019a_Iodine}) sample a state vector (using the reparameterization trick) and run this through the decoder to obtain, for each object, a predicted sub-image $\v{o}^{(k)}$ and segmentation logits $\v{\hat{m}}^{(k)}$. These decoder outputs are then fed, separately for each object, into the refinement subnetwork, along with additional inputs. Image-sized inputs are fed into the bottom convolutional layer, while vector-sized inputs are appended to the output of the final convolutional layer and processed by the final LSTM and Linear layers. We use the same selection of auxiliary inputs as~\citep{Greff2019a_Iodine}, except that we omit the leave-one-out likelihoods. 

Inference across multiple video frames is performed in a sliding window. This window slides "into" and "out of" the full video or episode. That is, the first window includes only the first video frame, which is processed for $L$ iterations, before moving the inference window ahead by one frame. Thus, it takes $L\times F$ inference steps before the window reaches its full extent, comprising $F$ frames. The reverse happens when we reach the end of an episode or video. The lagging end of the window keeps advancing until there is only a single frame left within the window, which gets processed for a final $L$ iterations. Thus, each video frame gets processed for exactly $L\times F$ inference steps.

\subsection{Training procedure}\label{sec:training-procedure}
The above network architecture was trained on pre-generated experience with the active dSprites environment. The main training set, used to train the world model, comprised 50,000 videos of 4 frames each. An additional validation set of 10,000 videos was sampled identically and independently to the training set. Each video was generated as follows. First, an instance of the active dSprites environment was randomly initialized, with three objects. Object shapes (square, ellipse or heart) were drawn uniformly. Colors were uniformly sampled from 5 evenly spaced values, independently for the R, G and B channels (a uniform grayscale background color was sampled from the same 5 intensity values). Orientations were sampled uniformly from the interval $[0, 2\pi]$. Object sizes were sampled uniformly from $[\frac{1}{6}], \frac{1}{3}]$, where $1$ is the size of the image frame. Positions were sampled uniformly from the interval $[0.2, 0.8]^2$ (where $0$ and $1$ are the edges of the frame), while velocities were drawn from a Normal distribution with mean $0$ and standard deviation $0.0625$. Video frames were then generated by simulating 4 frames of these objects' dynamics. Between the 2nd and 3rd frames, a single action (acceleration) was randomly sampled for each object, from a Normal distribution with mean $0$ and s.d. $0.0625$, and placed at a random pixel location within the object's segmentation mask. If an object happened to be completely invisible (e.g. because it was fully occluded by another), then it received no action. An additional action was sampled for a random background pixel, to encourage the model to learn to accurately segment the background into a separate object slot (rather than grouping the background with one of the objects). Training with these 4-frame videos was followed by a brief period of training with longer videos (12 frames, with actions after every second frame) that were otherwise identically sampled.

Additional sets of 50,000 training videos (one for each task; as well as accompanying sets of 10,000 identically and independently sampled validation videos) were used to train OBR's preference network. Each of these videos was 8 frames long and generated as follows. First, an instance of the active dSPrites environment was randomly initialized as before (but with a different procedure for sampling shapes -- see below). Then, 6 frames of this environment were simulated with random actions inserted before the 3rd and 5th frames. To prevent objects from leaving the frame (which is likely in longer videos and becomes problematic in this setting), we additionally placed actions as needed to prevent this from happening (specifically, if an object was on course to move to a coordinate outside the range $[0.05, 0.95]^2$ in the next frame, an action was generated that resulted in the object appearing to "bounce" against an invisible wall instead). Before the 7th and 8th frames, actions were generated that brought the objects to their target positions and velocities within the context of a task (see below). The first of these actions brought the objects to their target positions, while the second decelerated them to 0 velocity. Thus, in the final frame (and only then), the objects reached their target configuration. Each frame was encoded (using the normal inference procedure) as a set of latent beliefs in OBR's latent space, and the preference network was then tasked to map the representation of the first 7 "seed frames" to that of the final target frame, by minimizing the following KL-divergence loss:
\begin{gather}
    \mathcal{L_\text{pref}} = \sum_{t=1}^{T-1} \sum_k D_\text{KL}\left( \tilde{p}_t\left(\genstate^{(k)}\right) || q\left(\genstate^{(k)}_{T}\right) \right) \\
    \tilde{p}_t\left(\genstate^{(k)}\right) = \mathcal{N}\left(\genstate^{(k)}; \tilde{\v{\mu}}^{(k)}(\v{\lambda}_t ), {\tilde{\v{\sigma}}^{(k)}}(\v{\lambda}_t )^2  \right)
\end{gather}
where $\v{\lambda}_t$ are the parameters of the variational beliefs for frame $t$, and we make explicit the fact that the parameters (means and variances) of the predicted preference distribution are a function (via the preference network) of these beliefs about prior frames. 

Object target positions depended on the task. Specifically, the vertical target coordinates of any square, ellipse and heart shapes in the scene were fixed to 0.2, 0.5 and 0.8, respectively. If, under the task, objects were to go to the left of the frame, their target horizontal coordinate was 0.2 -- if they were to go to the right, it was 0.8. Object shapes were pseudo-randomized to evenly sample task conditions. In the \textit{IfHeart} task, we ensured that training episodes had a 50\% probability of containing at least one heart. In the \textit{HeartXORSquare} task, we sampled four conditions with equal probability: (1) there being neither a heart or a square in the scene; (2) at least one heart but no squares present; (3) at least one square but no hearts present; and (4) at least one heart and one square present in the scene. Outside of these restrictions, shapes were sampled randomly.

Training was performed using the ADAM optimizer \citep{kingma2014adam} with default parameters and an initial learning rate of $3\times10^{-4}$. This learning rate was reduced automatically by a factor 3 whenever the validation loss had not decreased in the last 10 training epochs, down to a minimum learning rate of $3\times10^{-5}$. Training was performed with a batch size of 64 (16 $\times$ 4 GPUs). OBR's world model was deemed to have converged after 241 epochs, which required approximately 24 hours to train on 4 Nvidia A100 GPUs. Preference networks for the \textit{IfHeart} and \textit{HeartXORSquare} tasks were trained for 90 and 120 epochs, respectively (18 and 24 hours).

\subsubsection{ELBO loss}
OBR optimizes the following (weighted) ELBO loss for both learning and inference:
\begin{multline}
    \mathcal{L} = -\sum_{t=0}^{T} \Bigg[ 
    \underbrace{\mathcal{H}\left(q\left(\{\v{s}^{\dagger(k)}_t, \v{a}^{(k)}_t\}\right) \right)}_\text{Complexity}
    + \underbrace{\beta E_{q(\{\v{s}^{(k)}_t\})}[\log p(\v{o}_t|\{ \v{s}^{(k)}_t\})]}_\text{Reconstruction accuracy}  \\
    + \underbrace{\sum_k E_{q(\v{a}^{(k)}_t)}[\log p(\v{a}^{(k)}_t|\v{\Psi}_t)]}_\text{Action inference accuracy}
    + \underbrace{ \sum_k E_{q\left(\v{s}^{\dagger(k)}_{t}, \v{s}^{\dagger(k)}_{t-1}, \v{a}^{(k)}_{t-1}\right)}[\log p(\v{s}^{\dagger(k)}_{t}|\v{s}^{\dagger(k)}_{t-1}, \v{a}^{(k)}_{t-1})]}_\text{Temporal consistency}  \Bigg]
\end{multline}
where $\mathcal{H}(\bullet)$ denotes entropy. Similar to previous work (e.g. \citep{Greff2019a_Iodine}) we up-weight the reconstruction accuracy term in this loss by a factor $\beta$. We train the network to minimize not just the loss at the end of the inference iterations through the network, but a composite loss that also includes the loss after earlier iterations. Let $\mathcal{L}_\beta^{(n)}$ be the loss after $n$ inference iterations, then the composite loss is given by:
\begin{equation}
    \mathcal{L}_\text{comp}= \sum_{n=1}^{N_\text{iter}} \frac{n}{N_\text{iter}}\mathcal{L}^{(n)}
\end{equation}

\subsubsection{Hyperparameters}
OBR includes a total of 4 hyperparameters: (1) the loss-reweighting coefficient $\beta$ (see above); (2) the variance of the pixels around their predicted values, $\sigma_o^2$; (3) the variance of the noise in the latent space dynamics, $\sigma_s^2$; and (4) the variance of the noise in the object actions, $\sigma_\psi^2$. The results described in the current work were achieved with the following settings:
\\ \\
\noindent\begin{tabular}{cc}
\toprule
\bfseries Param.~~ & \bfseries Value~~ \\
\midrule
$\beta$       & 5.0 \\
$\sigma_o$    & 0.3 \\
$\sigma_s$    & 0.1 \\
$\sigma_\psi$ & 0.3 \\
\bottomrule
\end{tabular}

\section{Computing $E_{ q(\mathbf{a}^{(k)} ) } [ \log p(\mathbf{a}^{(k)}|\mathbf{\Psi}) ] $} 
\label{sec:samplingtrick}
The expectation under $q(\v{a}^{(k)})$ of $\log p(\v{a}^{(k)}|\v{\Psi})$, which appears in the ELBO loss (eq. \ref{eq:elbo}), cannot be computed in closed form, because the latter log probability requires us to marginalize over all possible configurations of the pixel-to-object assignments, and to do so inside of the logarithm. That is:
\begin{align}
    \log p(\v{a}^{(k)}|\v{\Psi}) &=  \log \left( \sum_\v{\mask} p(\v{a}^{(k)}|\v{\Psi}, \v{\mask}) p(\v{\mask}|\{ \v{s}^{(k)} \}) \right) \\
    &= \log \left( E_{p(\v{\mask}|\{ \v{s}^{(k)} \})} [ p(\v{a}^{(k)}|\v{\Psi}, \v{\mask}) ] \right)
\end{align}
However, note that within the ELBO loss, we want to maximize the expected value of this quantity (as its negative appears in the ELBO, which we want to minimize). From Jensen's inequality, we have:
\begin{equation} \label{eq:jensen}
      E_{p(\v{\mask}|\{ \v{s}^{(k)} \})} [ \log p(\v{a}^{(k)}|\v{\Psi}, \v{\mask}) ] 
    \leq \log \left( E_{p(\v{\mask}|\{ \v{s}^{(k)} \})} [ p(\v{a}^{(k)}|\v{\Psi}, \v{\mask}) ] \right)
\end{equation}
Therefore, the l.h.s. of this equation provides a lower bound on the quantity we want to maximize. Thus, we can approximate our goal by maximizing this lower bound instead. This is convenient, because this lower bound, and its expectation under $q(\v{a}^{(k)})$ can be approximated through sampling:
\begin{gather}
     E_{q(\v{a}^{(k)})} \left[ E_{p(\v{\mask}|\{ \v{s}^{(k)} \})} [ \log p(\v{a}^{(k)}|\v{\Psi}, \v{\mask}) ] \right] \approx \frac{1}{N_\text{samples}} \sum_j  \log p(\v{a}^{(k)*}_j|\v{\Psi}, \v{\mask}^*_j) \\
     = \frac{1}{N_\text{samples}} \sum_j \log \mathcal{N}\left(\v{a}^{(k)*}_j; \sum_i \hat{\mask}^{*(i)}_{jk}\v{\psi}_i, \sigma_\psi^2\v{I}\right) \\
     \v{\hat{\mask}}^{*(i)}_j \sim  p(\mask_i|\{ \v{s}^{(k)} \}), \quad
     \v{a}^{(k)*}_j \sim q(\v{a}^{(k)}), \quad 
     \v{s}^{(k)*}_j \sim q(\v{s}^{(k)})
\end{gather}
where we slightly abuse notation in the sampling of the pixel assignments, as a vector is sampled from a distribution over a categorical variable. The reason this results in a vector is because this sampling step uses the Gumbel-Softmax trick~\citep{Jang2016gumbel}, which is a differentiable method for sampling categorical variables as "approximately one-hot" vectors. Thus, for every pixel $i$, we sample a vector $\v{\hat{\mask}}^{*(i)}_j$, such that the $k$-th entry of this vector, $\hat{\mask}^{*(i)}_{jk}$, denotes the "soft-binary" condition of whether pixel $i$ belongs to object $k$. In practice, we use $N_\text{samples}=1$, based on the intuition that this will still yield a good approximation over many training instances, and that we rely on the refinement network to learn to infer good beliefs. The Gumbel-Softmax sampling method depends on a temperature $\tau$, which we gradually reduce across training epochs, so that the samples gradually better approximate the ideal one-hot vectors. 

It is worth noting that, as the entropy of $p(\v{\mask}|\{ \v{s}^{(k)} \})$ decreases (i.e. as object slots "become more certain" about which pixels are theirs), the bound in equation \ref{eq:jensen} becomes tighter. In the limit as the entropy goes to 0, the network is perfectly certain about the pixel assignments, and so the distribution collapses to a point mass. The expectation then becomes trivial, and so the two sides of eq.~\ref{eq:jensen} become equal. Sampling the pixel assignments is equally trivial in this case, as the distribution has collapsed to permit only a single value for each assignment. In short, at this extreme point, the procedure becomes entirely deterministic. In our data, we typically observe very low entropy for $p(\v{\mask}|\{ \v{s}^{(k)} \})$, and so we likely operate in a regime close to the deterministic one, where the approximation is very accurate. 

\subsection{Planning}
\label{sec:planning-details}
OBR's linearized dynamics model allows us to find an optimal action plan $\v{\pi}^*$ (minimizing equation~\ref{eq:planning}) in closed form, as follows:
\begin{gather}
    \v{d}^{(T)} = \begin{bmatrix} 1 \\ 2 \\ \vdots \\ T \end{bmatrix}, \quad \v{\Omega}_d^{(T)} = \begin{bmatrix} 
    1 & 0 & \hdots & 0 & 0 \\
    1 & 0 & \hdots & 0 & 0 \\ 
    2 & 1 & \hdots & 0 & 0 \\ 
    1 & 1 & \hdots & 0 & 0 \\ 
    \vdots & \vdots & \ddots & \vdots & \vdots \\ 
    T-1 & T-2 & \hdots & 1 & 0 \\ 
    1 & 1 & \hdots & 1 & 0 \\ 
    T & T-1 & \hdots & 2 & 1  \\ 
    1 & 1 & \hdots & 1 & 1 \end{bmatrix} \\ 
    \v{U}=(\v{\Omega}_d^{(T)} \otimes \v{D}), \quad \v{L} = \text{diag}(\v{1}\otimes \tilde{\v{\sigma}})^{-2} \label{eq:Dmap} \\
     \v{\pi}^* = (\v{U}^T \v{L} \v{U} + \lambda_a \v{I})^{-1} \v{U}^T \v{L} \left( \v{1} \otimes (\tilde{\v{\mu}} - \v{\mu}_{s^\dagger_t}   ) - \v{d}^{(T)} \otimes \begin{bmatrix} \v{\mu}_{s'_{t}} \\ 0 \end{bmatrix} \right)
\end{gather}
where $\v{1}$ denotes a column vector of 1s of length $T$, $\otimes$ denotes the Kronecker product, and $\v{\mu}_{s^\dagger_t}$ and $\v{\mu}_{s'_{t}}$ denote, respectively, the means of the variational beliefs about the full object state in generalized coordinates ($\genstate$), and the means of the beliefs about the derivatives ($\statederiv$) only. In words, $\v{U}$ is a matrix that maps actions within the planning horizon to their (cumulative) effects over that time window. These effects are translated to the network's latent space, through the multiplication by $\v{D}$ in equation~\ref{eq:Dmap}. The optimal actions are obtained by projecting the current (precision-weighted) error (i.e., the discrepancy between the desired state and the sequence of states that will unfold if no action is taken) onto the (precision-weighted) pseudoinverse of $\v{U}$. This projection is optionally shrunk towards a zero-mean Gaussian prior over actions, with precision $\lambda_a$, to regularize the scale of the actions. 

In the experiments reported in this work, we used a planning horizon of $T=3$ and an action regularization strength $\lambda_a=0.1$. We also set $\v{L}=\v{I}$ as we empirically find this leads to better and more stable performance.

\subsubsection{Baseline configurations}

 All ablation experiments and baseline comparisons were conducted on a single NVIDIA H200 (141 GB) with PyTorch 2.7 and CUDA 12.8. SAVi variants were trained from scratch using Adam optimizer (lr = $1 \times 10^{-4}$ with cosine decay to $1 \times 10^{-6}$) on 50k Active-dSprites sequences with 4 slots (128-dim). V-JEPA variants share a frozen ViT-L encoder (1024-dim), 12-layer predictor (512-dim), and action head (12-dim). Frame stacking (1–8 frames) and learning rates ($10^{-3}$–$10^{-6}$) were systematically explored. Optimal configurations: V-JEPA Original (4 frames, L1 + rollout loss), V-JEPA BC (1 frame, MSE loss), V-JEPA DAgger (2 frames, MSE loss with iterative data aggregation, $\beta = 0.8$ decay). Dynamics module ablations (linear, GRU, GateL0RD, LSTM, Transformer) were trained on 8k state trajectories with batch sizes optimized per architecture. 
 
 \textbf{Training duration.} SAVi variants: 50 epochs. 1-frame variant approximately 1 hour; 4-frame variant required approximately 3.5 hours, 8-frame variant  approximately 6.5 hours; and 16-frame variant required approximately 12 hours. V-JEPA 2-AC Original: 100 epochs (6 hours), did not converge. V-JEPA 2-AC BC: 50 epochs (3.2 hours), converged at epoch 26 (1.5 hours). V-JEPA 2-AC DAgger: 50 epochs across 6 data aggregation iterations (6 hours total), converged at epoch 22 (2 hours). Dynamics modules: 50 epochs (1–3 hours per architecture depending on model complexity).

\section{Latent space traversals}
\label{sec:appendix:traversals}
OBR is trained without supervision, and so the structure of its latent representational space is \textit{a priori} unknown. Here, we explore this space by systematically traversing it one dimension at a time. We first let OBR encode an image into its latent space. Subsequently, for each object representation, we increment the target dimension's encoded (mean) value with a range of 20 evenly spaced values between $[-1, 1]$. We then feed the resulting, perturbed latent vectors through OBR's decoder to obtain 20 images that vary systematically (and for all objects) along the target dimension. The images in Fig. \ref{fig:latents} show the results of this procedure for three different scenes, for each of the 16 latent dimensions of the trained OBR model. Of particular note are dimensions 3 and 9, which have learned to encode the objects' positions (along an approximately orthogonal set of axes at an arbitrary angle to the pixel grid). Other recognizable variations can be seen in color (dimensions 4, 6 and 11) and size (14). Shape and orientation appear to be entangled along multiple dimensions. Other dimensions do not obviously encode anything -- they may be truly non-coding, or their role might not be visible in combination with the other latent values that we happened to sample here. In particular, the depth value of an object is almost certainly encoded in one or more latent dimensions, but this is not apparent from these traversals, as they target all objects simultaneously, and thus should not affect their occlusion patterns (which object is in front of which other object(s)).
\newpage
\vspace{-1cm}
\begin{figure}[hbtp!]
	\centering
    \begin{subfigure}{1\textwidth}
    \includegraphics[width=\textwidth]{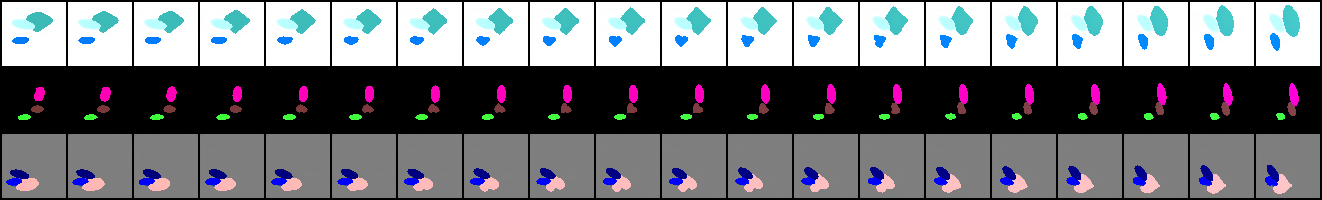}
    \caption{\textbf{Latent dimension 0}}     
    \end{subfigure}
    \begin{subfigure}{1\textwidth}
    \includegraphics[width=\textwidth]{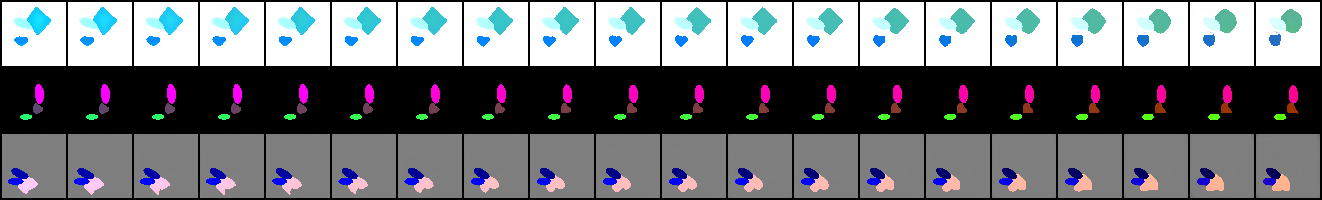}
    \caption{\textbf{Latent dimension 1}} 
    \end{subfigure}
    \begin{subfigure}{1\textwidth}
    \includegraphics[width=\textwidth]{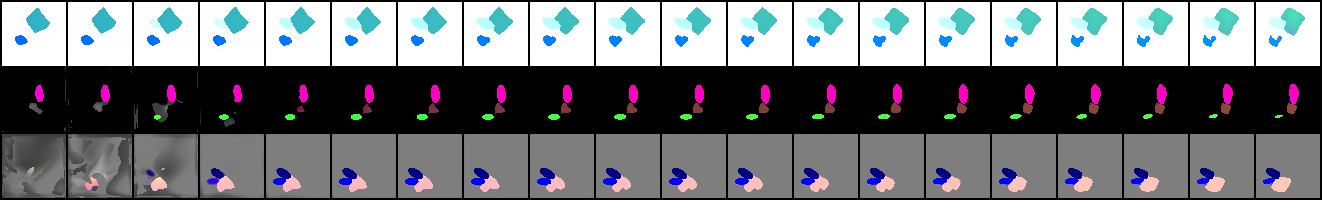}
    \caption{\textbf{Latent dimension 2}} 
    \end{subfigure}
    \begin{subfigure}{1\textwidth}
    \includegraphics[width=\textwidth]{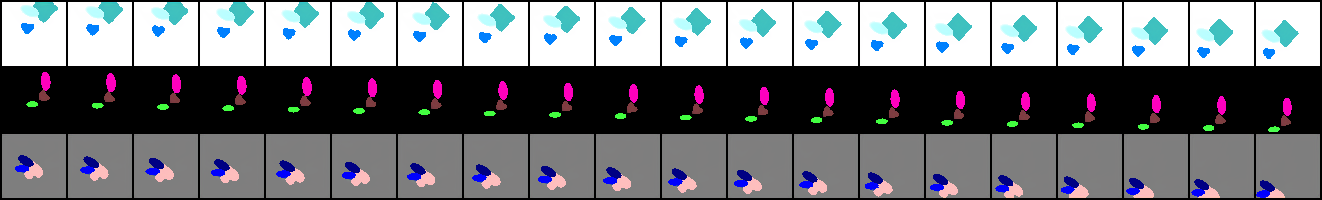}
    \caption{\textbf{Latent dimension 3}} 
    \end{subfigure}
    \begin{subfigure}{1\textwidth}
    \includegraphics[width=\textwidth]{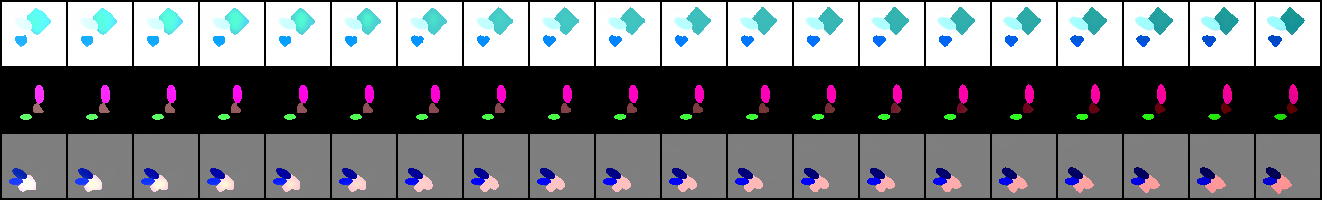}
    \caption{\textbf{Latent dimension 4}} 
    \end{subfigure}
    \begin{subfigure}{1\textwidth}
    \includegraphics[width=\textwidth]{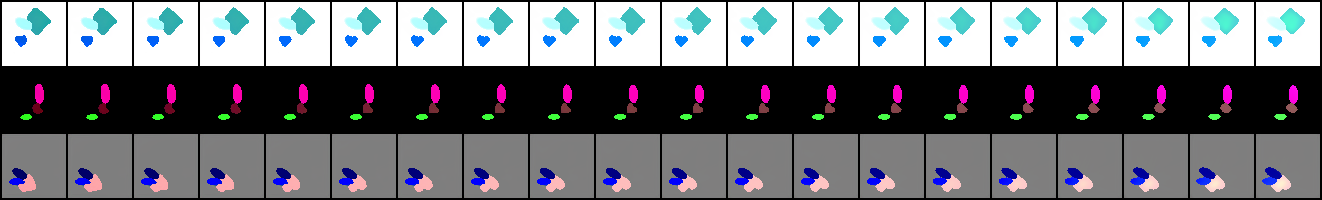}
    \caption{\textbf{Latent dimension 5}} 
    \end{subfigure}
    \begin{subfigure}{1\textwidth}
    \includegraphics[width=\textwidth]{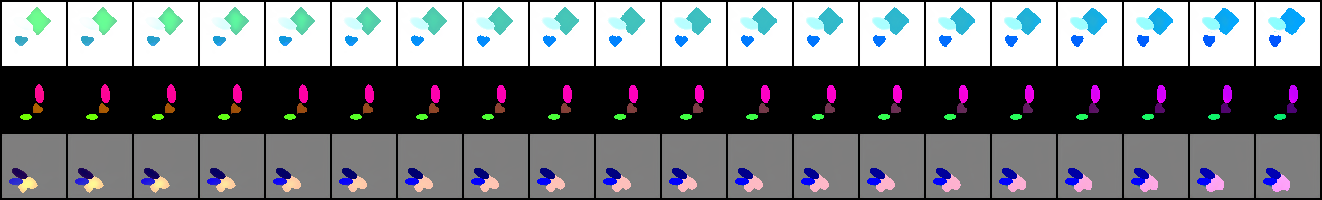}
    \caption{\textbf{Latent dimension 6}} 
    \end{subfigure}
    \begin{subfigure}{1\textwidth}
    \includegraphics[width=\textwidth]{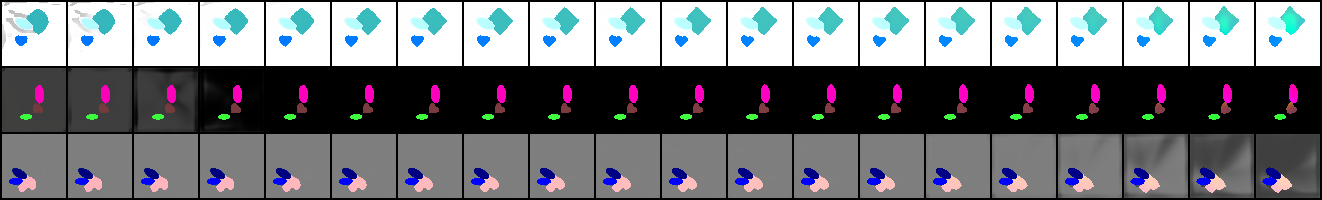}
    \caption{\textbf{Latent dimension 7}} 
    \end{subfigure}
    \caption{} \label{fig:latents}
\end{figure}
\begin{figure}[hbtp!]
	\centering    
    \begin{subfigure}{1\textwidth}
    \includegraphics[width=\textwidth]{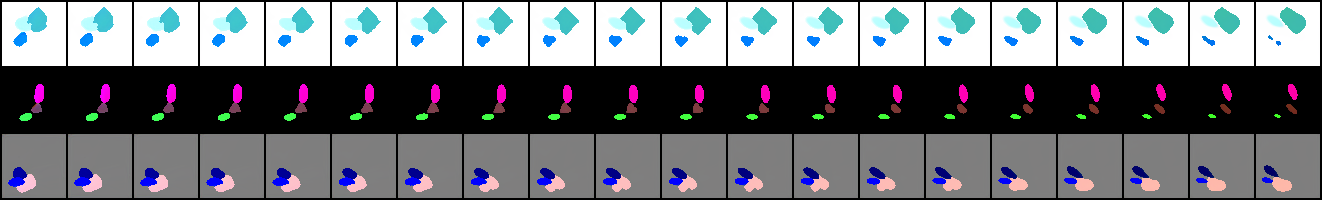}
    \caption{\textbf{Latent dimension 8}} 
    \end{subfigure}
    \begin{subfigure}{1\textwidth}
    \includegraphics[width=\textwidth]{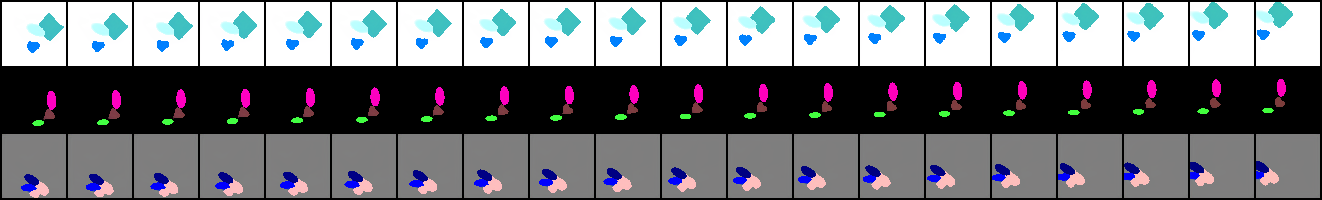}
    \caption{\textbf{Latent dimension 9}} 
    \end{subfigure}
    \begin{subfigure}{1\textwidth}
    \includegraphics[width=\textwidth]{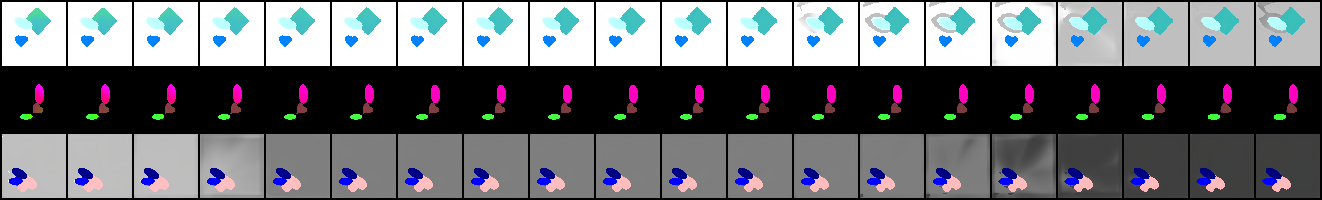}
    \caption{\textbf{Latent dimension 10}} 
    \end{subfigure}
    \begin{subfigure}{1\textwidth}
    \includegraphics[width=\textwidth]{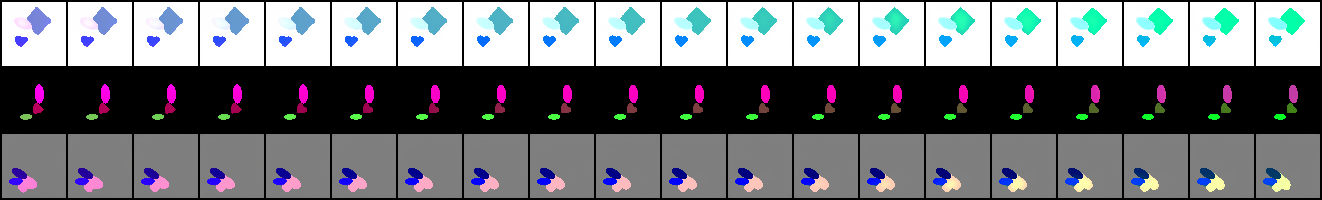}
    \caption{\textbf{Latent dimension 11}} 
    \end{subfigure}
    \begin{subfigure}{1\textwidth}
    \includegraphics[width=\textwidth]{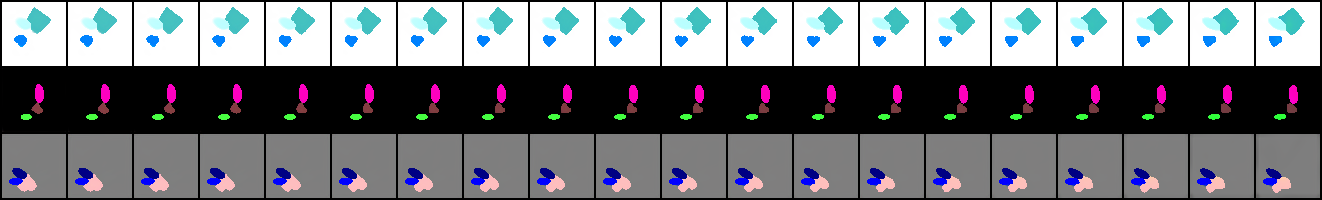}
    \caption{\textbf{Latent dimension 12}} 
    \end{subfigure}
    \begin{subfigure}{1\textwidth}
    \includegraphics[width=\textwidth]{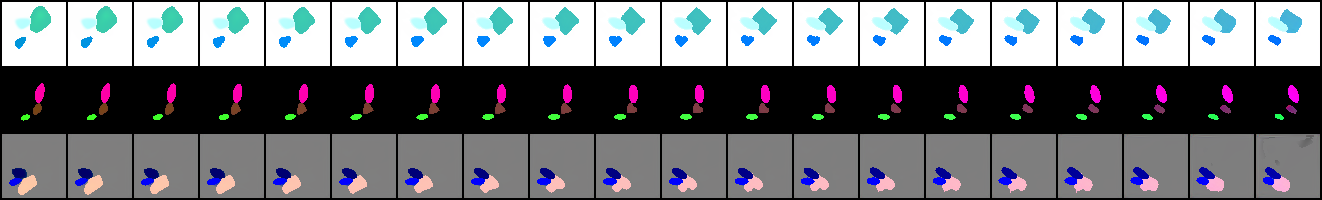}
    \caption{\textbf{Latent dimension 13}} 
    \end{subfigure}
    \begin{subfigure}{1\textwidth}
    \includegraphics[width=\textwidth]{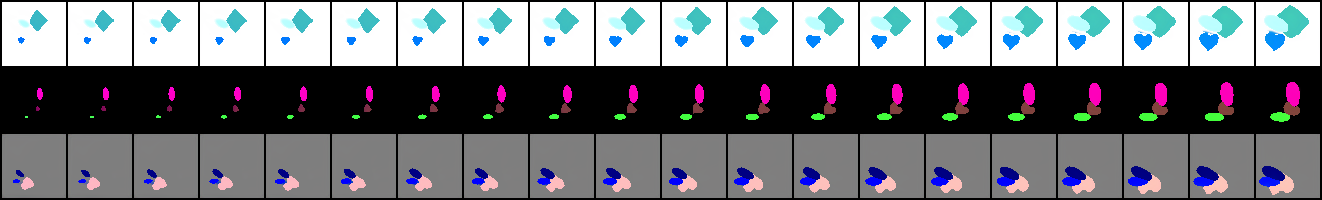}
    \caption{\textbf{Latent dimension 14}} 
    \end{subfigure}
    \begin{subfigure}{1\textwidth}
    \includegraphics[width=\textwidth]{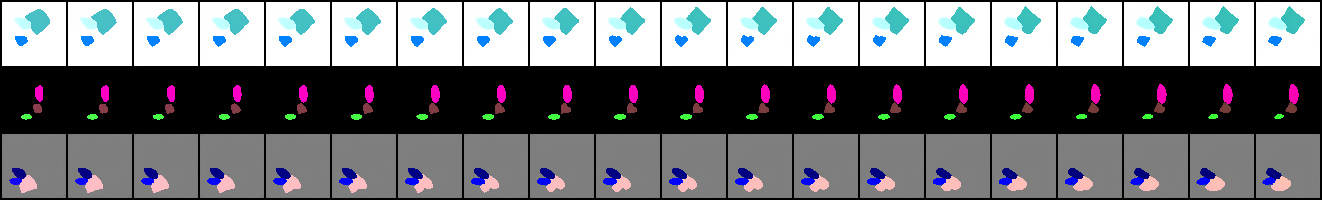}
    \caption{\textbf{Latent dimension 15}} 
    \end{subfigure}
    \label{fig:latents1}
\end{figure}

\newpage

 \bibliographystyle{elsarticle-num} 
 \bibliography{OBR}

\end{document}